\documentclass[acmlarge]{acmart}

\setcopyright{acmlicensed}
\copyrightyear{2018}
\acmYear{2018}
\acmDOI{XXXXXXX.XXXXXXX}

\acmJournal{TOIS}

\usepackage{multirow}%
\usepackage[caption=false,font=footnotesize,labelfont=rm,textfont=rm]{subfig}

\begin{document}

\title{Multi-Focused Video Group Activities Hashing}

\author{Zhongmiao Qi}
\email{876789574@qq.com}

\affiliation{%
  \institution{Ningbo University}
  \city{Ningbo}
  \state{Zhejiang}
  \country{China}
}

\author{Yan Jiang}
\email{2403567035@qq.com}
\affiliation{%
	\institution{Ningbo University}
	\city{Ningbo}
	\state{Zhejiang}
	\country{China}
}

\author{Bolin Zhang}
\email{zhangbolin@nbu.ebu.cn}
\affiliation{%
  \institution{Ningbo University}
  \city{Ningbo}
  \state{Zhejiang}
  \country{China}
}

\author{Chong Wang}
\email{wangchong@nbu.edu.cn}
\affiliation{%
 \institution{Ningbo University}
 \city{Ningbo}
 \state{Zhejiang}
 \country{China}}

\author{Lijun Guo}
\email{guolijun@nbu.ebu.cn}
\affiliation{%
	\institution{Ningbo University}
	\city{Ningbo}
	\state{Zhejiang}
	\country{China}
}

\author{Pengjiang Qian}
\email{qianpjiang@jiangnan.edu.cn}
\affiliation{%
  \institution{Jiangnan University}
  \city{Wuxi}
  \state{Jiangsu}
  \country{China}}

\author{Jiangbo Qian}
\email{qianjiangbo@nbu.edu.cn}
\authornote{Corresponding Author. Faculty of Electrical Engineering and Computer Science, Ningbo University;
Merchants' Guild Economics and Cultural Intelligent Computing Laboratory, Ningbo University}
\authornotemark[0]
\affiliation{%
 \institution{Ningbo University}
 \city{Ningbo}
 \state{Zhejiang}
 \country{China}}

\renewcommand{\shortauthors}{Qi et al.}

\begin{abstract}
With the explosive growth of video data in various complex scenarios, quickly retrieving group activities has become an urgent problem. However, many tasks can only retrieve videos focusing on an entire video, not the activity granularity. To solve this problem, we propose a new STVH (spatiotemporal interleaved video hashing) technique for the first time. Through a unified framework, the STVH simultaneously models individual object dynamics and group interactions, capturing the spatiotemporal evolution on both group visual features and positional features. Moreover, in real-life video retrieval scenarios, it may sometimes require activity features, while at other times, it may require visual features of objects. We then further propose a novel M-STVH (multi-focused spatiotemporal video hashing) as an enhanced version to handle this difficult task. The advanced method incorporates hierarchical feature integration through multi-focused representation learning, allowing the model to jointly focus on activity semantics features and object visual features. We conducted comparative experiments on publicly available datasets, and both STVH and M-STVH can achieve excellent results.
\end{abstract}


\ccsdesc[500]{ Information systems ~Top-k retrieval in databases}

\keywords{Group activity recognition, Video understanding, Video group activity retrieval, Hash learning}

\received{20 February 2007}
\received[revised]{12 March 2009}
\received[accepted]{5 June 2009}

\maketitle

\section{Introduction}\label{sec1}

Video group activity, which consists of interactions among multiple objects, is the core of video analysis because it can express high-level semantic information. For example, robbery activities can alert automatically for surveillance monitors \cite{real1, Social}, kills in volleyball matches might be searched for score calculation \cite{VD, RWGCN}, and goals in football games are replayed as they are exciting moments \cite{soccer1,soccer2}. With the explosive growth of video data in various complex scenarios, finding and retrieving activities quickly has become an urgent task.

\begin{figure}[]
	\centering  
	\subfloat[Existing video hashing methods gene\\-rate hash codes from a global perspective.]{
		\label{fig1_a}\includegraphics[width=0.3\textwidth]{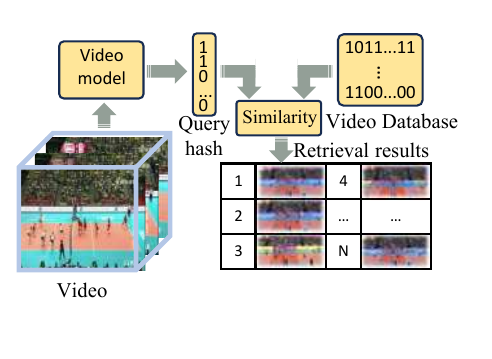}
	}
	\subfloat[Existing group activities classification methods, which directly and simply fuses positional and visual features to activity classification.]{
		\label{fig1_b}\includegraphics[width=0.3\textwidth]{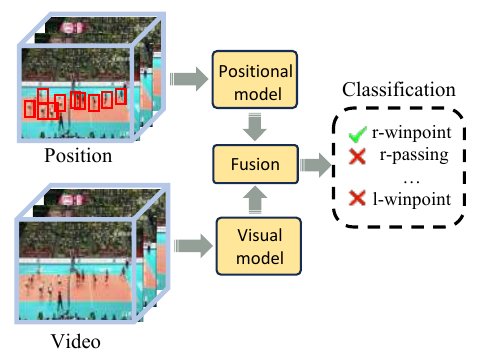}
	} \\
	\subfloat[Our video activity hashing method interleaves visual features with positional features for activity hash coding and activity classification.]{
		\label{fig1_c}\includegraphics[width=0.6\textwidth]{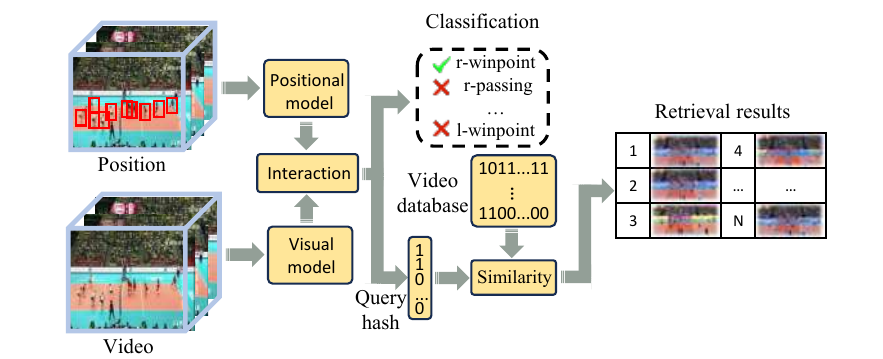}
	}\\  
	\subfloat[Ours multi-focused video hashing methods, the multi-focused hash codes  are achieved by progressively filtering out positional change information at each layer using a binary filtering matrix.]{
		\label{fig1_d}\includegraphics[width=0.6\textwidth]{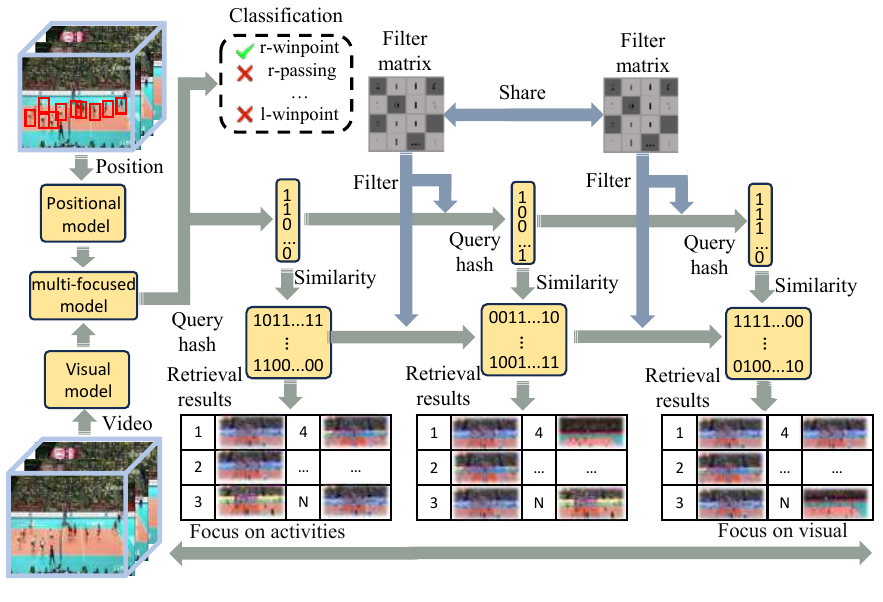}
	}          
	
	\caption{Our methods and existing methods.}   
	\label{fig1}
\end{figure}

Hash learning \cite{hash1,hash2,hashing, image_hash} is a high-speed and efficient technique widely used in large-scale data retrieval, which can keep the distance between hash codes consistent with the distance between the original data. As shown in Fig. \ref{fig1_a}, existing video hashing methods encode videos from a global perspective rather than activity-focused encoding. Some activity-focused methods (non-hash methods) can only perform categorization \cite{AFGf, MLST,CAD}, which cannot satisfy the speed requirement when the data volume is large. As shown in Fig. \ref{fig1_b}, these methods only perform simple fusion operations (e.g., add or concat) of visual and positional features, which are unable to effectively perform activity modeling. Furthermore, in real-life video retrieval scenarios, sometimes it may need to emphasize activities (i.e., activity-focused hash), while other times it may need to emphasize visual features of objects (i.e., visual-focused hash). Take football match videos as an example, we may retrieve offensive segments that emphasize activities, or we may retrieve segments only from one team that emphasize visual features. Therefore, if only one set of hash codes can meet the above requirements, it can significantly reduce storage cost.

To address these challenges, we propose a spatiotemporal video hashing (STVH) technique, as illustrated in Fig. \ref{fig1_c}. STVH models group activities in videos by jointly capturing visual and positional changes of both individual objects and groups, thereby generating compact hash codes. Furthermore, to generate different focus hash codes, we extend STVH to a multi-focused spatiotemporal video hashing method, which utilizes a multi-step fusion module to aggregate visual features and location features gradually. Additionally, we introduce a binary filtering matrix to refine positional features in the hash code, enhancing its sensitivity to the visual information. The key contributions of our work are summarized as follows:

\begin{itemize}
	\item[$\bullet$]To accelerate the retrieval of similar videos at an activity semantic granularity, we propose a new STVH  technique for the first time. A novel positional and visual features deep fusion (PVF) module can interleave and fuse object visual features with position features.
	
	\item[$\bullet$]To meet the requirements of activity-focused hash codes or visual-focused hash codes by only one set of codes, we further propose a new M-STVH technique, with a binary filtering matrix to reduce storage cost. 
	
	\item[$\bullet$]A contrastive learning loss based on object interrelationships is proposed to maintain the distance between hash codes of similar activities.
	
	\item[$\bullet$]The experimental results show that the STVH and M-STVH achieve competitive results in classification accuracy on multiple group activity recognition datasets, even when using hash codes. Excellent performance is also demonstrated for either visual-focused hash codes or activity-focused hash codes.
\end{itemize}

The rest of this paper as follows: Section \ref{sec2} describes the related work. Section \ref{sec3} introduces the problem definition and symbol definition. Section \ref{sec4} introduces the details of the SVTH model. Section \ref{sec5} introduces the details of the M-STVH improvement. Section \ref{sec6} shows the experimental results of the STVH method and a comparison with other methods. Section \ref{sec7} summarizes this work and provides an outlook for future research.

\section{Related Work}\label{sec2}
To our knowledge, this paper is the first work to propose the activity hash retrieval problem. Here, we briefly introduce two similar tasks: video hash retrieval and activity recognition.

\subsection{Video Hashing}
Video hash retrieval methods have predominantly based on deep learning, which can be classified into supervised and unsupervised hash learning approaches. Supervised hash learning generally extracts video features via a neural network, followed by a hash code generation method through label-based supervision. SPTDH \cite{SPTDH} introduced a similarity-preserving deep temporal hashing network that unifies video modeling and hash learning into a single, cohesive process. DVSH \cite{DVSH} employed a 3D convolution\cite{3D} extracted spatiotemporal features from videos, subsequently mapping the extracted features into a unified binary space. BrVAE \cite{BrVAE} enabled an uncertainty-aware video hashing by predicting the probability distribution of hash codes, thus offering robust uncertainty quantification. 

Unsupervised video hash learning methods initially relied on autoencoders or clustering-based approaches for model training. SSVH \cite{SSVH} designed a hierarchical binary autoencoder, where a convolutional neural network serves as the video encoder and an LSTM \cite{LSTM} acts as the decoder, allowing the model to learn video features across multiple dimensions. TSVH \cite{TSVH} employed a transformer-based autoencoding network combined with time-sensitivity regularization, effectively minimizing sensitivity to local temporal disturbances while retaining global temporal sequence information. By utilizing a hash-based affinity matrix, this approach effectively preserves pairwise similarity between video samples. Recently, with the advent of contrastive learning techniques, some researchers have extended this unsupervised learning paradigm to video hash learning. COMH \cite{COMH} generated two distinct views of the video features through two different random masking techniques, then leverages contrastive learning to maximize the similarity of the hash codes between the views, followed by hash learning via video reconstruction from the hash codes. Dns \cite{Dns} built a re-ranking framework based on a knowledge distillation scheme and a selection mechanism that allows large unlabeled datasets to train our network of students and selectors. It not only improves the efficiency of retrieval but also ensures competitive performance. KPSC-P and KPSC-F \cite{KPSC} utilized pre-trained visual language models as knowledge sources while prioritizing video functionality to enhance diversity and reduce noise, ultimately achieving unsupervised video retrieval. 

\subsection{Group Activity Recognition}
Group Activity recognition methods can be categorized into traditional and deep learning-based approaches, depending on the use of deep learning techniques. Traditional group behavior recognition predominantly depends on handcrafted feature extraction (e.g., HOG \cite{HOG}, SIFT \cite{SIFT}), followed by the use of probabilistic graphical models like POMDPs or AND-OR logic for behavior prediction. Recently, with the rapid advancement of deep learning, architectures such as ResNet \cite{ResNet}, LSTM \cite{LSTM}, GCN \cite{GCN}, and GRU \cite{GRU} have made significant strides in the field of group behavior recognition. Several state-of-the-art group activity recognition methods are predominantly based on deep neural network models. For instance, MOGAR \cite{MOGAR} organically combined joint motion, trajectories, and object positions to generate richer activity representations, further enhancing the corresponding features via gating mechanisms and self-attention mechanisms, leading to the final group behavior classification. AFGNet \cite{AFGf} introduced a third-order active factor graph network that simulates the third-order interactions between every triplet of active object. To enhance group consistency, it incorporates a consistency-aware inference module, which includes two penalty terms that model the inconsistency between object and group activity. \cite{STAHs} investigated spatial and temporal correlations, proposing a novel loss function for self-supervised group action recognition. RWGCN \cite{RWGCN} introduced a random walk graph convolutional network for group activity recognition, incorporating a Levy flight random walk mechanism within GCN to capture information from various nodes, while leveraging prior positional information to recognize group activities. ASTFormer \cite{ASTFormer} utilized CNN to extract image features and then design an action-centered aggregation strategy, grouping objects performing different actions before making predictions. DIN \cite{DIN} introduced a spatiotemporal dynamic reasoning model that predicts the relational matrix and captures dynamic walking offsets through a joint processing method that integrates dynamic relation and dynamic walking modules. MLST-Former \cite{MLST} presented a multi-level spatiotemporal Transformer-based relational reasoning framework aimed at exploring temporal dependencies and spatial dynamics among different objects in group activities.

\subsection{Summary}
Although video content is effective in conveying semantic information, current approaches encounter significant challenges in efficiently retrieving both online and historical activities. Furthermore, existing video-based hashing methods primarily operate on the entire granularity videos, lacking the capability to encode activities at the video semantic information, which substantially restricts their applicability in real-world scenarios. In addition, current approaches to group activity classification tend to focus solely on activity features, neglecting crucial information related to the participating entities. In practice, visual information about the objects involved plays a pivotal role in comprehensively understanding the video.

\begin{figure*}[t]
	
	\centerline{\includegraphics[width=16cm]{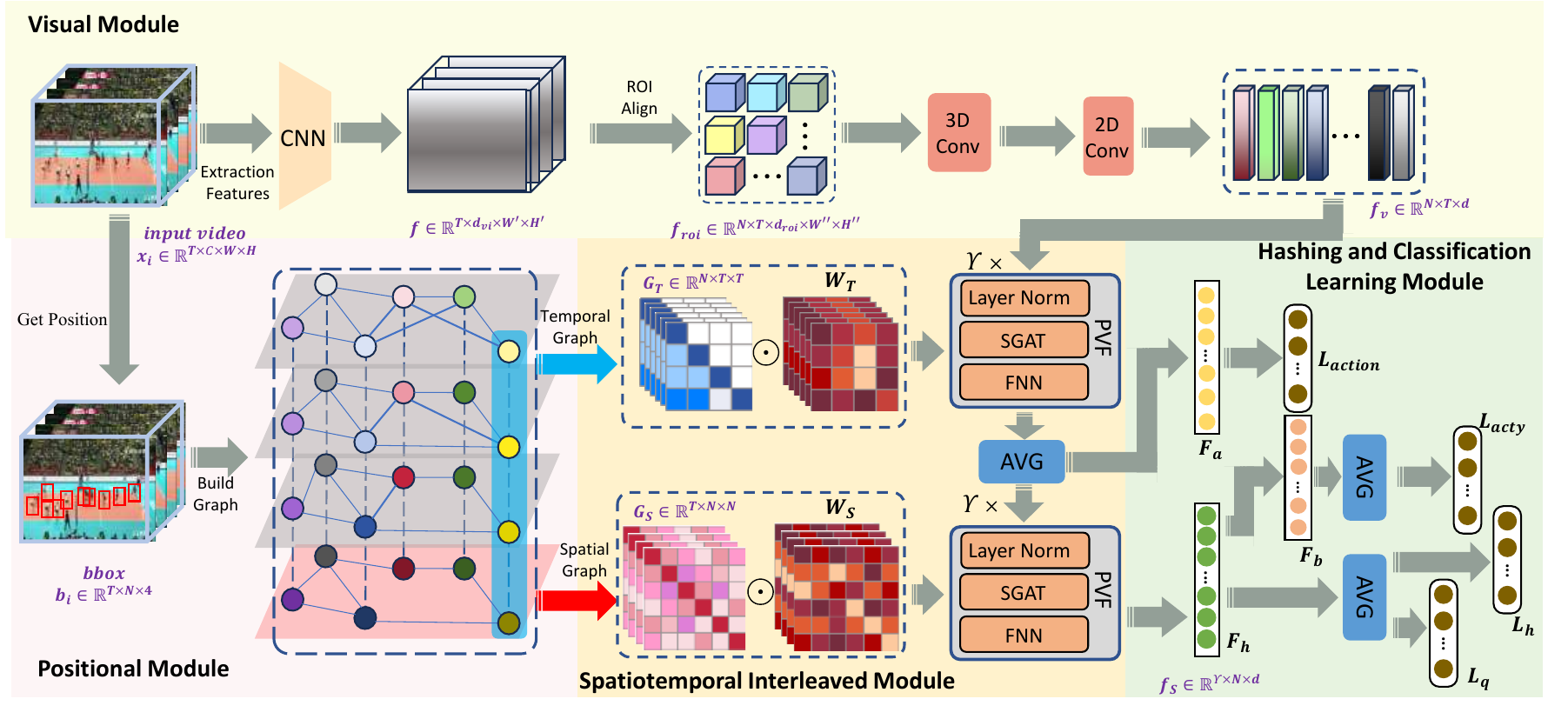}}
	\caption{STVH consists of four main modules: 1) Visual Module: extracting features from the input video; 2) Positional Module: modeling spatiotemporal relationships on the input position information; 3) Spatiotemporal Interleaving Module: interleaving visual features and positional features; and 4) Hashing and Classification Learning Module: fusing the visual features with the position features and outputting the corresponding hash codes and classifications.}
	\label{fig_2}
\end{figure*}

\section{Problem Definition}\label{sec3}
Assume the training set input video is $\boldsymbol{X}=\{\boldsymbol{x}_1,\boldsymbol{x}_2,...,\boldsymbol{x}_M\}$ and the position information is $\boldsymbol{Boxes}=\{\boldsymbol{boxes}_1,\boldsymbol{boxes}_2, \\... ,\boldsymbol{boxes}_M \}$, which has a activity class label of $\boldsymbol{Y}^{acty}=\{\boldsymbol{y}^{acty}_1,\boldsymbol{y}^{acty}_2,... ,\boldsymbol{y}^{acty}_M \}$, where $M$ denotes the number of video samples. $\boldsymbol {x}_i \in \mathbb{R}^{T\times C \times W\times H}$ denotes that the input video has $T$ frames, and the scale of each frame is $C \times W\times H$. $\boldsymbol{boxes}_i\in \mathbb{R}^{T\times N \times 4}$ is the position information of $N$ objects in video $\boldsymbol{x}_i$, and $\boldsymbol{box}_{j_t}$ is the position information of the $j$-th object in video $\boldsymbol{x}_i$ on the $t$-th frame. $\boldsymbol{Y}^{action}=\{\boldsymbol{y}^{action}_{i1},\boldsymbol{y}^{action}_{i2},... ,\boldsymbol{y}^{action}_{iN} \}$ denotes the object action in video $\boldsymbol {x}_i$. The objective of STVH is to map the input video and location information into a $K$-bits hash code while M-STVH is to map these inputs into multiple $K$-bits hash code with a different focus. The specific notation description is shown in Table \ref{tab:1}.

\begin{table*}[]
	\centering
	\caption{Explanatory Table of Selected Symbols}
	\label{tab:1}
	\begin{tabular}{cc}
		\hline
		Symbol                                     & Definition                                                  \\ \hline
		$\boldsymbol{x}_i$                         & Video $i$                                                     \\
		$\boldsymbol{boxes}_i$                     & Position of objects in video $i$                             \\
		$\boldsymbol{y}_i^{acty}$                  & Activity classification of video $i$                          \\
		$\boldsymbol{y}_{ij}^{action}$             & Action classification of object $j$ in video $i$                \\
		$\boldsymbol{b}_i$                         & Hash codes generated from video $i$                          \\
		$\widetilde{\boldsymbol{y}}_{i}^{acty}$    & Predicting the activity classification of video $i$           \\
		$\widetilde{\boldsymbol{y}}_{ij}^{action}$ & Predicting the action classification of object $j$ in video $i$ \\
		$T$                                        & Number of video frames                                      \\
		$M$                                        & Number of videos                                            \\
		$N$                                        & Number of objects in a video                                \\
		$K$                                        & Length of hash codes                                        \\
		$\boldsymbol{F}$                           & Filter matrix                                               \\
		$\boldsymbol{W}$                           & Learnable weight matrix                                     \\
		$\boldsymbol{G}_T$                         & Temporal relationship graph                                 \\
		$\boldsymbol{G}_S$                         & Spatial relationship graph                                  \\
		$B$                                       & Batch size \\ \hline
	\end{tabular}
\end{table*}

\section{STVH Model}\label{sec4}
The STVH in Fig. \ref{fig_2} is comprised of visual, positional, spatiotemporal interleaving, and classification hash learning modules.
\subsection{Visual Module}\label{subsec1}
The vision module extracts visual features from each object in the input video $\boldsymbol{x}_i$. First, we perform multiscale features extraction on the video frames using an ImageNet-pretrained Inception-v3 model (\cite{inv3}), yielding a feature tensor ${\boldsymbol{f} \in \mathbb {R}^{T\times d_v\times W'\times H' }}$ where $d_v=512$ and $W'$ and $H'$ are the scales of the feature map after downsampling $W$ and $H$ 32 times. We then extract object visual features $\boldsymbol{f}_{roi}\in \mathbb{R}^{N\times T \times d_v\times W''\times H'' }$ by their corresponding positional $\boldsymbol{boxes}_i$ via RoIAlign, where  $W''$ and $H''$ are both 5. To compactly represent these features, we apply sequential 3D and 2D convolutions to vectorize $\boldsymbol{f}_{roi}$, producing a unified visual feature ${\boldsymbol{f}_{v} \in \mathbb{R}^{N\times T\times d}}$  with a dimension of $d=1024$.

\subsection{Positional Module}\label{subsec2}

Although visual features are important in inferring group activities, relying solely on object features still poses challenges. For example, ``running" and ``jogging" may exhibit nearly identical visual patterns, making them difficult to distinguish without additional cues. To address this, we analyze the spatiotemporal positional changes of objects, which provide critical discriminative signals for activity recognition. Specifically, our module takes the positional coordinates of all objects in the video as input and models their trajectories to capture activity-specific motion dynamics.

\begin{figure}[]
	\centerline{\includegraphics[width=0.5\columnwidth]{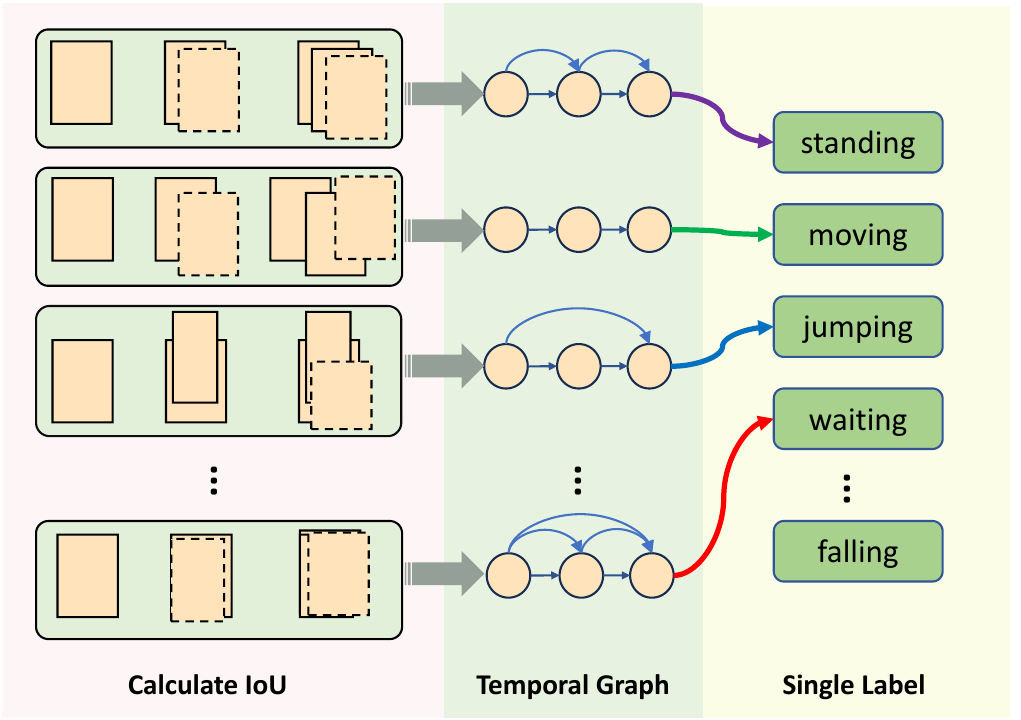}}
	\caption{ Modeling the action based on the computational IoU.}
	\label{fig_3}
\end{figure}

As shown in Fig. \ref{fig_3}, the Intersection over Union (IoU) between objects in consecutive frames effectively captures motion features, serving as both a speed indicator and trajectory estimator. A rapidly moving object exhibits low inter-frame IoU values, while slower movement produces higher IoU. By tracking IoU trends across multiple frames, we can infer motion patterns: consistently decreasing IoU suggests sustained unidirectional movement, while increasing IoU indicates oscillatory behavior like round-trip motion. This makes frame-wise IoU computation a robust quantitative measure for analyzing object movement dynamics. The IoU calculation formula is as follows:

\begin{equation}
	\label{IoU}\operatorname{IoU}\left(j, t_1, t_2\right) = \frac{\boldsymbol{box}_{j_{t_1}} \cap \boldsymbol{box}_{j_{t_2}}}{\boldsymbol{box}_{j_{t_1}} \cup \boldsymbol{box}_{j_{t_2}}},
\end{equation}
where $\boldsymbol{box}_{j_{t_1}}$ and $\boldsymbol{box}_{j_{t_2}}$ denote the position information of object $j$ at frames $t_1$ and $t_2$, respectively. We model temporal dependencies through directed edges, where only preceding frames influence subsequent ones. The final temporal graph $\boldsymbol{G}_T\in R^{N\times T\times T}$ is generated by applying a learnable weight matrix $\boldsymbol{W}$ to the IoU-based adjacency matrix. This adaptive weighting mechanism enables the network to dynamically adjust inter-node connections, while the directed graph topology effectively captures temporal object interactions throughout the video.

The spatial relation graph captures inter-object interactions by modeling proximity-based correlations. Consistent with the natural intuition that nearby objects tend to interact more strongly, we compute pairwise spatial relationships using normalized Euclidean distances between object positions. These relationships are encoded in a spatial graph $\boldsymbol{G}_S \in \mathbb{R}^{T\times N \times N}$, where the edge weights represent the strength of spatial dependencies between objects at each timestep. This graph structure enables explicit modeling of distance-based object interactions throughout the video clip. The normalized Euclidean distances calculation formula is as follows:
\begin{equation}
	\label{dist}
	\boldsymbol{d i s t}_{i j}=1-\sqrt{\frac{\left(\boldsymbol { box }_{i}-\boldsymbol { box }_{j}\right)^{2}}{\boldsymbol{std}}},
\end{equation}
where $\boldsymbol{std} \in \mathbb{R}^{T \times 4}$ is the standard deviation of multiple dimensions and $\boldsymbol{d i s t}_{i j}$ is the normalized Euclidean distance between two objects in multiple time frames. We directly take $\boldsymbol{d i s t}_{i j}$ as the corresponding element in $\boldsymbol{G}_S$.

\subsection{Spatiotemporal Interleaving Module}

Since group activities are composed of interactions among multiple objects in a video, integrating the spatiotemporal information of these objects becomes crucial. STVH integrates two granularities of spatiotemporal interleaving, thereby obtaining representations of group activities. First, it models object actions based on changes in visual features and positional features of objects in videos. Second, it models group activities based on changes in group visual features and their positional features. Additionally, visual and positional features complement each other in modeling actions and activities. Therefore, we propose a position and visual deep fusion Module (PVF) to integrate positional and visual features.

\begin{figure}[]
	\centerline{\includegraphics[width=0.5\columnwidth]{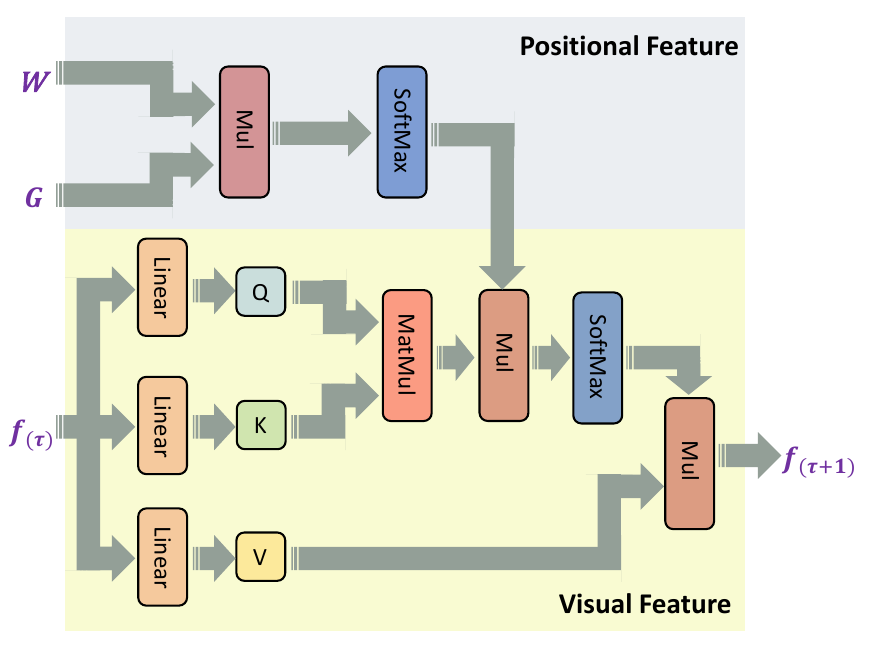}}
	\caption{Sparse graph relation attention module (SGAT), fusing visual features as well as positional features at an attention.}
	\label{SRAT}
\end{figure}

The spatiotemporal interleaving operations shown in Fig. \ref{fig_2} are implemented through sequential stacking $\mathit{\Upsilon }$ times of PVF modules. Each PVF module consists of three components: a normalization layer, a sparse graph relation attention (SGAT, Fig. \ref{SRAT}) layer, and feedforward neural networks (FFNs). During the iterative PVF processing, high-level semantic information from positional features becomes progressively fused with visual features. The operational flow begins with input visual features $\boldsymbol{f}_{({\tau})}$ undergoing layer normalization to produce $\boldsymbol{f}'$. These normalized features $\boldsymbol{f}'_{({\tau})}$ are then combined with the spatiotemporal graph $\boldsymbol{G}$ through the SGAT module to achieve fusion of positional dynamics and visual features, resulting in $\boldsymbol{f}''_{({\tau})}$. The transformed features $\boldsymbol{f}''_{({\tau})}$ subsequently pass through the FNN for nonlinear mapping, ultimately generating the output features $\boldsymbol{f}_{({\tau}+1)}$ for the current iteration. This process can be formally described as follows:

\begin{equation}
	\begin{array}{l}
		\boldsymbol{f}'_{(\tau) } = \text{LayerNorm}(\boldsymbol{f}_{(\tau) }), \\
		\\
		\boldsymbol{f}''_{(\tau)} = \text{SGAT}(\boldsymbol{f}'_{(\tau)}, \boldsymbol{G}), \\
		\\
		\boldsymbol{f}_{(\tau +1)} = \text{FNN}(\boldsymbol{f}''_{(\tau)}),
	\end{array}
\end{equation}
where $\tau$ denotes the $\tau$-th layer ($1<=i<=\mathit{\Upsilon }$). Additionally, as described in the previous section, the temporal graph $\boldsymbol{G}_T$ and the spatial graph $\boldsymbol{G}_T$ represent the positional features of objects in the video. Therefore, to interleave positional features with visual features, we employ the SGAT module to achieve multi-granularity interleaving. Fig. \ref{SRAT} illustrates the SGAT calculates attention matrices for positional features and visual features separately, then integrates them to obtain the final attention matrix. We introduce a trainable parameter matrix $\boldsymbol{W}$ to accommodate the floating-point values in the graph $\boldsymbol{G}$. We perform a nonlinear transformation on these values and then generate the position feature attention matrix via a softmax operation, specifically the dot product of $\boldsymbol{G}$ and $\boldsymbol{W}$. These matrices are then multiplied by the attention matrix derived from visual features to obtain the final attention matrix. The SRAT computation process is as follows:
\begin{equation}
	\begin{array}{l}
		\operatorname{SGAT}(\boldsymbol{f}, \boldsymbol{G})=\boldsymbol{W}_{1} \boldsymbol{f} \text { Attention }(\boldsymbol{f}, \boldsymbol{G}), \\
		\\
		\text { Attention }(\boldsymbol{f}, \boldsymbol{G})=\operatorname{AT}_{v}(\boldsymbol{f}) \times \operatorname{AT}_{p}(\boldsymbol{G}), \\
		\\
		\operatorname{AT}_{v}(\boldsymbol{f})=\operatorname{SoftMax}\left(\frac{\left(\boldsymbol{W}_{2} \boldsymbol{f}\right)\left(\boldsymbol{W}_{3} \boldsymbol{f}\right)^{T}}{\sqrt{C}}\right), \\
		\\
		\operatorname{AT}_{p}(\boldsymbol{G})=\operatorname{SoftMax}(\boldsymbol{G} \times \boldsymbol{W}),
	\end{array}
\end{equation}
where $\boldsymbol{W}_1$, $\boldsymbol{W}_2$, and $\boldsymbol{W}_3$ are three same shape learnable weight matrices. $\text{AT}_v()$ computes the visual attention matrix, and $\text{AT}_p()$ computes the positional attention matrix. The temporal feature $\boldsymbol{f}_T$ and the group feature $\boldsymbol{f}_S$ of the activity are obtained after fusion.

\subsection{Hashing and Classification Learning Module}
After obtaining the $\boldsymbol{f}_S$ and $\boldsymbol{f}_T$,  hash learning and classification of activities can be performed. The features $\boldsymbol{f}_T$ are first mean-pooled in on the time dimension and then fed into the fully connected layer $\boldsymbol{F}_a$, thereby obtaining the action class $\widetilde{\boldsymbol{y}}_{ij}^{action}$ of each object. 

We input $\boldsymbol{f}_S$ into the fully connected layer $\boldsymbol{F}_h$ and normalize it to obtain the floating-point value encoding $\boldsymbol{h}_i$, which is then binarized to obtain the video hash code $\boldsymbol{b}_i$. We then input $\boldsymbol{h}_i$ into the fully connected layer $\boldsymbol{F}_c$ to get the group activity prediction results $\widetilde{\boldsymbol{y}}_{i}^{acty}$ for the video. The operations can be written as:

\begin{equation}
	\begin{array}{l}
		\boldsymbol{h}_i = \text{sign}(\text{AVG}(\boldsymbol{F}_h(\boldsymbol{f}_S))), \\
		\\
		\widetilde{\boldsymbol{y}}_{i}^{acty} = \boldsymbol{F}_c(\boldsymbol{h}_i),
	\end{array}
\end{equation}

\subsection{Loss Function}
STVH employs a loss function that consists of classification loss $L_{cls}$, hash loss $L_{q}$, and hash contrastive loss $L_{CON}$. The $L_{cls}$ is composed of action classification loss $L_{\text{action}}$ and activity classification loss $L_{\text{acty}}$. The hash loss uses quantization loss. Hash contrastive loss is based on the relationships between multiple objects.

The classification loss enables the model to obtain feature representations of specific activities classified during training. We use a cross-entropy loss for classification constraints, which can be defined as:

\begin{equation}
	\begin{array}{c}
		L_{\text {acty }}=\sum_{i=1}^{M}-\boldsymbol{y}_{i}^{\text {acty }} \log \left(\widetilde{\boldsymbol{y}}_{i}^{{acty }}\right), \\
		L_{\text {action }}=\sum_{i=1}^{M} \sum_{j=1}^{N}-\boldsymbol{y}_{i j}^{\text {action }} \log \left(\widetilde{\boldsymbol{y}}_{i j}^{{action }}\right).
	\end{array}
\end{equation}

The STVH generates floating-point values during training, while our objective is to output binary hash codes. Consequently, there is information loss when converting a float to binary. The exponential contrastive loss can reduce the loss incurred during conversion by minimizing the discrepancy between binary codes and float values. The equation can be written as:

\begin{equation}
	L_q = \sum_{i=1}^{M} \sum_{j=1}^{M} exp(\frac{1}{K} |\boldsymbol{h}_i^T \boldsymbol{h}_j - \boldsymbol{b}_i^T \boldsymbol{b}_j|),
\end{equation}
where $\boldsymbol{h}_i^T \boldsymbol{h}_j$ is the inner product of the float values and $\boldsymbol{b}_i^T \boldsymbol{b}_j$ is the binarized inner product.

Since group activities emerge from interactions among multiple objects, activities with similar interaction patterns among objects should be alike. Therefore, we enhance the contrastive loss by considering the relationship between various objects to maintain the interclass distance for different activities. Specifically, we utilize the predicted action labels $\widetilde{\boldsymbol{y}}_{i j}^{{action }}$ as node features and construct a spatial relationship graph $\boldsymbol{G}_S$ adjacency matrix. This matrix is then input into a graph convolutional network (GCN) to model interactions among multiple objects, resulting in encoded representations $\boldsymbol{a}_i$.

Therefore, $\boldsymbol{a}_i$ and $\boldsymbol{b}_i$ represent the encoding of group activities in the same video using two different methods. We can consider $\boldsymbol{b}_i$ as a more detailed encoding of group activities generated through all the information in the video, while $\boldsymbol{a}_i$ is generated based on the spatial relationships between multiple objects, representing a relatively fuzzy encoding of group activities. Since they originate from the same video, they should exhibit similarity. In this paper, we employ a contrastive loss based on associations between objects to minimize the difference between $\boldsymbol{a}_i$ and $\boldsymbol{b}_i$, thus ensuring that the model maintains proximity between similar activities even when trained using only class label supervision. This can be written as:

\begin{equation}
	\begin{array}{c}
		L_{con(i, i)}=\sum_{i=1}^{B}\log \frac{\sum_{j=1}^{B} \left({sim}\left(\boldsymbol{a}_{i}, \boldsymbol{b}_{j}\right)\right)+ \left({sim}\left(\boldsymbol{a}_{j}, \boldsymbol{b}_{i}\right)\right)}{{sim}\left(\boldsymbol{a}_{i}, \boldsymbol{b}_{i}\right)}, \\
		\\
		{sim}\left(\boldsymbol{a}_{i}, \boldsymbol{b}_{{i}}\right)=\exp(\frac{\boldsymbol{a}_{i}^{T} \boldsymbol{b}_{i}}{\left\|\boldsymbol{a}_{i}\right\|_{2}\left\|\boldsymbol{b}_{i}\right\|_{2}}) , \\
		\\
		L_{CON}=\sum_{i=1}^{B} L_{con(i, i)} ,
	\end{array}
\end{equation}

\subsection{Optimization}
According to the above discussion, the loss function for model training mainly consists of four losses: $L_{acty}$, $L_{action}$, $L_q$ and $L_{CON}$. $L_{acty}$ and $L_{action}$ constitute $L_{cls}$, and $L_{cls}$=$L_{acty}$+0.5$L_{action}$. The total loss can be written as:
\begin{equation}
	L = L_{cls} + \lambda_1 L_{q} + \lambda_2 L_{CON},
\end{equation}
where $\lambda_1$ and $\lambda_2$ are hyperparameters to balance the losses, which must be obtained according to the actual retrieval effect in the training process.

\begin{figure*}[h]
	
	\centerline{\includegraphics[width=16cm]{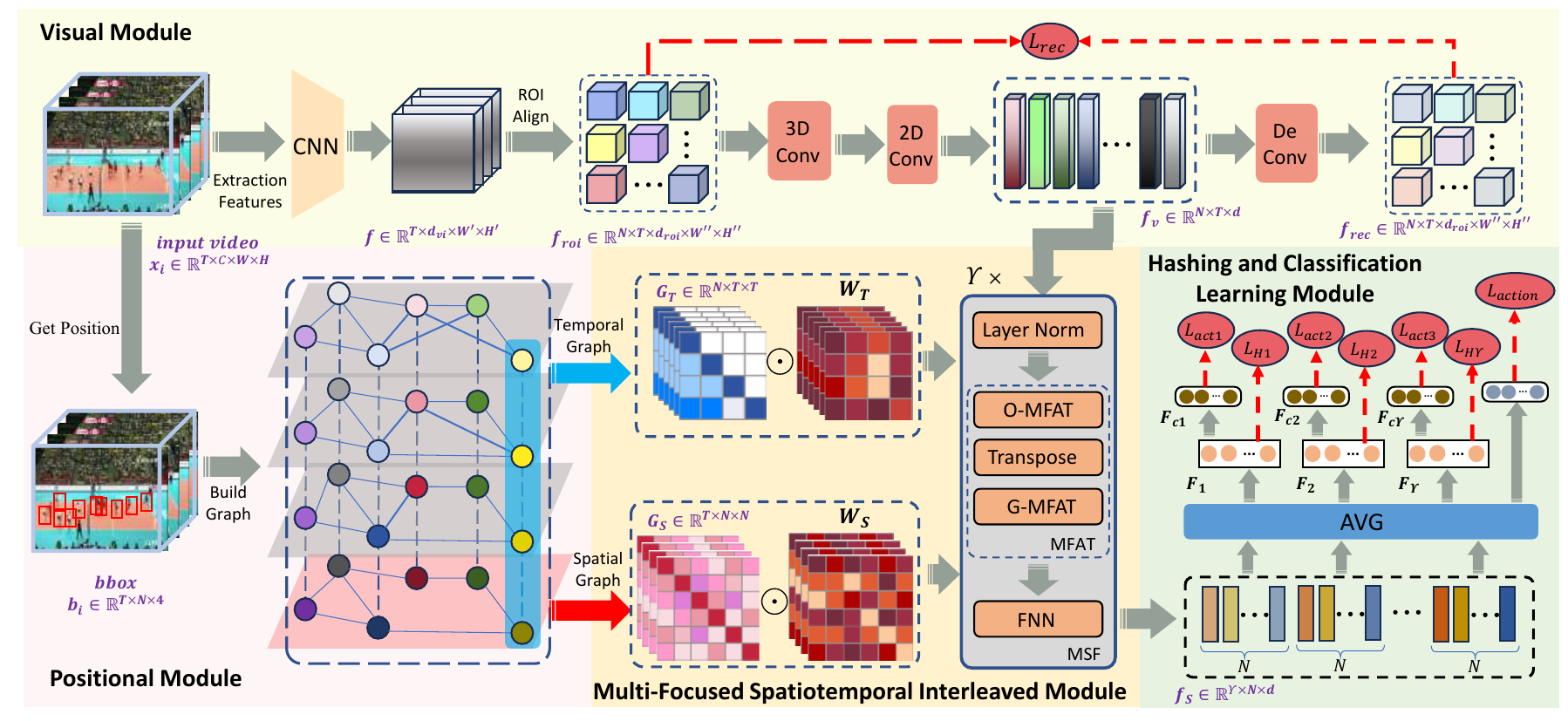}}
	
	\caption{M-STVH are composed of four modules: 1) Visual Module: extracting features from the input video; 2) Positional Module: modeling spatiotemporal relationships on the input position information; 3) Multi-Focused Spatiotemporal Interleaving Module: interleaving visual features and positional features at multiple layers; and 4) Hashing and Classification Learning Module: fusing the visual features with the position features and outputs the corresponding hash codes and classifications.}
	\label{M-STVH}
\end{figure*}

\section{M-STVH}\label{sec5}
While the STVH model effectively generates activity hash codes for video group activities, real-world retrieval scenarios often require more flexible representations. Applications may demand either activity-focused hashes (emphasizing group activity) or visual-focused hashes (emphasizing object visual information). To this end, we further improve the STVH model to perform the difficult task.

The M-STVH model architecture, illustrated in Fig. \ref{M-STVH}, comprises four key components: a visual module, a positional module, a multi-focused spatiotemporal interleaved module, and a hashing and classification learning module. While maintaining structural similarities to STVH in its visual and positional modules, M-STVH introduces unsupervised feature reconstruction to enhance discriminative visual feature extraction. The multi-focued spatiotemporal interleaving module achieves comprehensive modeling through multi-focued processing. The module gradually integrates the object positional features with the object visual features, allowing the extracted features to transition smoothly from a visual-focused to an activity-focused approach. Finally, the hash and classification learning module simultaneously converts multi-focus features into or focus hash codes and activities classifications.

\subsection{Multi-Focused Spatiotemporal Interleaved Module}
To comprehensively capture multi-focused group features in videos, we explore methods for integrating visual and positional features across multiple layers of objects and groups. As shown in Fig. 5, our model is mainly composed of stacking $\mathit{\Upsilon }$ times through MSF modules. Within each MSF module, there is a layer normalization module, a multi-fusion attention module (MFAT), and a feedforward neural network module. As the number of MSF stacking increases, the model incorporates more positional features into the visual features, allowing for a dynamic shift from visual to positional feature. By selecting outputs at different MSF layers, we obtain features with different emphases: the outputs from shallower layers emphasize group visual features, while those from deeper layers emphasize group activities. The specific calculation process can be written as:
\begin{figure}[]
	\centerline{\includegraphics[width=0.8\columnwidth]{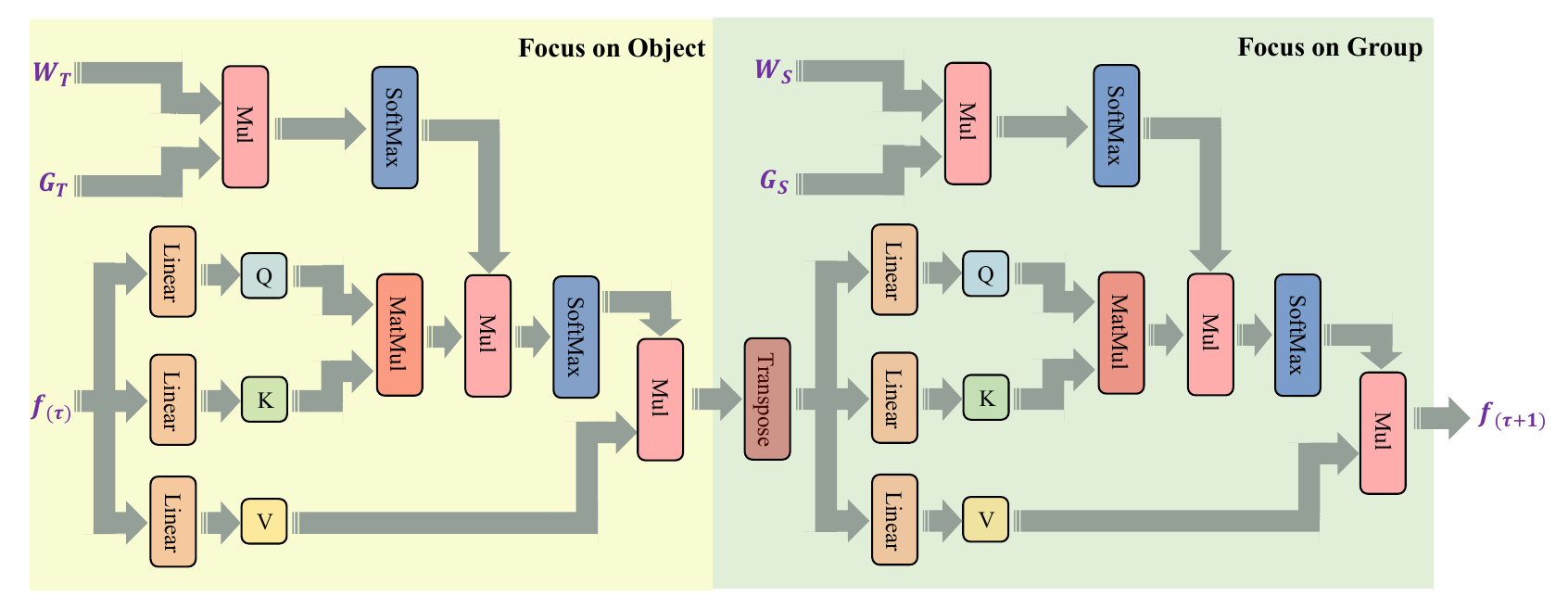}}
	\caption{Multi-fusion attention module, fusing visual features as well as positional features at object and group.}
	\label{MFAT}
\end{figure}

\begin{equation}
	\begin{array}{c}
		\boldsymbol{f}'_{(\tau)} = \text{LayerNorm}(\boldsymbol{f}_{(\tau)}), \\
		\\
		\boldsymbol{f}''_{(\tau)} = \text{MFAT}(\boldsymbol{f}'_{(\tau)},\boldsymbol{G}_T, \boldsymbol{G}_S), \\
		\\
		\boldsymbol{f}_{(\tau+1)} = \text{FFN}(\boldsymbol{f}''_{(\tau)}),
	\end{array}
\end{equation}
where $\tau$ denotes the $\tau$-th layer ($1<=\tau<=\mathit{\Upsilon }$). As mentioned in the previous section, the temporal relation graph $\boldsymbol{G}_T$ encodes the motion trajectory information of objects in the video sequence. In contrast, the spatial relation graph $\boldsymbol{G}_S$ characterizes the spatial interactions among objects within a frame. The core of the MSF module is the MFAT module, which is shown in Fig. 6. When the MFAT module focus the object (O-MFAT), it captures the action combining the temporal graph as well as the visual features of objects, and when the MFAT module focus on the group (G-MFAT), it understands the group activity by combining the spatial graph as well as the visual features of the group. Specifically, this process can be described as follows:
\begin{equation}
	\begin{array}{l}
		\text { MFAT }(\boldsymbol{f},\boldsymbol{G}_T, \boldsymbol{G}_S) =  \text{G-MFAT}(\text{O-MFAT}(\boldsymbol{f}, \boldsymbol{G}_T), \boldsymbol{G}_S), \\ 
		\\
		\text{G-MFAT}(\boldsymbol{f},\boldsymbol{G}_S) =  (\text{AT}_v(\boldsymbol{f})\times \text{AT}_p(\boldsymbol{G}_T))(\boldsymbol{W}_g \boldsymbol{f}), \\
		\\
		\text{O-MFAT}(\boldsymbol{f},\boldsymbol{G}_T) = (\text{AT}_v(\boldsymbol{f})\times \text{AT}_p(\boldsymbol{G}_T))(\boldsymbol{W}_o \boldsymbol{f}), \\
	\end{array}
\end{equation}
where $\text{AT}_v()$ and $\text{AT}_p()$ calculate as Eq. 4, and $\boldsymbol{W}_g$ and $\boldsymbol{W}_o$ are both learnable matrices. Through the processing of stack MSF modules, we can obtain multi-focused video features $ \left \{  \boldsymbol{f}_{(\tau)} \right \} _{(\tau=1)}^\mathit{\Upsilon} $ at each layer, forming a multi-focused video feature that spans from group visual to group activities. Specifically, in the shallow layer, the model focuses more on extracting static visual features of objects, resulting in shallow-layer features that represent the visual features of objects participating in a group activity.  When the network depth increases, deeper-layer features progressively incorporate richer positional and spatiotemporal interaction information, thereby emphasizing the dynamic features of group activity. 

\subsection{Hashing and Classification Learning Module}
After obtaining the multi-focused video features $ \left \{  \boldsymbol{f}_{(\tau)} \right \} _{(\tau=1)}^\mathit{\Upsilon} $, we proceed with hash learning and activity classification. First, we reduce the dimensionality of the features through a fully connected layer followed by a pooling operation to obtain $ \left \{  \boldsymbol{h}_{(\tau)} \right \} _{(\tau=1)}^\mathit{\Upsilon} $, and then binarize them using the sign function to generate multi-focused hash codes $ \left \{  \boldsymbol{b}_{(\tau)} \right \} _{(\tau=1)}^\mathit{\Upsilon} $. This process can be formally described as:
\begin{equation}
	\begin{array}{c}
		\boldsymbol{h}_{(\tau)} = \boldsymbol{F}_{(\tau)}(\text{AVG}(\boldsymbol{f}_{(\tau)})), \\
		\\ 
		\boldsymbol{b}_{(\tau)} = \text{sign}(\boldsymbol{h}_{(\tau)}),
	\end{array}
\end{equation}
where $1 \le \tau \le \mathit{\Upsilon}$. To obtain more accurate video hash code representations, we perform classification tasks for both group activities and individual actions on the features before binarization. Specifically, the real-valued features $ \left \{  \boldsymbol{h}_{(\tau)} \right \} _{(\tau=1)}^\mathit{\Upsilon} $ are fed into multiple distinct group activities classification heads to produce hierarchical group activity classification outputs $\left\{{\boldsymbol{\widetilde{y}}}_{\left(\tau\right)}^{acty}\right\}_{\left(\tau=1\right)}^\mathit{\Upsilon}$. For individual action classification, we utilize only the output features $   \boldsymbol{h}_{(\mathit{\Upsilon})} $ from the final layer to generate predictions for individual action categories $\boldsymbol{y}^{action}$. This approach ensures that the resulting hash codes encapsulate both comprehensive group activity semantics and detailed individual action features, enhancing their discriminative power for video retrieval tasks.   

\subsection{Loss Function}
Compared to the loss function in STVH, we added a reconstruction loss for visual features in M-STVH. The loss focuses on object features and uses mean squared error (MSE) loss. Specifically, we compute the MSE loss between the original $\boldsymbol{f}_{roi}$ and reconstructed features $\boldsymbol{\widetilde{f}}_{roi}$:
\begin{equation}
	L_{recon} = \frac{1}{d}(\boldsymbol{f}_{roi} - \boldsymbol{\widetilde{f}}_{roi})^2,
\end{equation}
where $d$ denotes the dimension of the feature.  To enable the model to learn discriminative feature representations for specific activities, we employ the cross-entropy loss function to constrain the classification tasks. Considering that features at different foci contain semantic information of varying granularity, we assign a hyperparameter as a weight to each prediction to regulate its contribution. Specifically, the classification loss weight for shallow-layer features is smaller, as these features primarily encode basic visual information. As the depth increases, the classification loss weight increases, since these features incorporate richer activity information. This hierarchical weighting strategy allows the model to adaptively balance feature learning across different layers, which can be defined as:
\begin{equation}
	L_{a c t y}=\sum_{i=1}^{r}-w_{(i)} \boldsymbol{y}^{\text {acty }} \log \left(\widetilde{\boldsymbol{y}}_{(i)}^{\text {acty }}\right),
\end{equation}
where $w_{\left(i\right)}$ represents a hyperparameter. For object actions, we also apply the cross-entropy loss as a constraint, and its loss computation is like Eq. 6.
The overall classification loss $L_{cls}=L_{acty}+0.5L_{action}$. Subsequently, both hash loss $L_q$ and hash contrastive loss  $L_H$ are similar to those in STVH, as shown in Subsection 4.5. The total loss can be written as:
\begin{equation}
	L_{total} = L_{cls} + \mu_1 L_q + \mu_2 L_H + \mu_3 L_{recon},
\end{equation}
where $\mu_1$, $\mu_2$ and $\mu_3$ are hyperparameters to balance the losses, which must be obtained according to the actual retrieval effect in the training process.

\subsection{Filter Matrix}
Our M-STVH can effectively generate multi-focused hash codes. However, multi-focused hash codes cost more storage space than a single hash code. To solve this problem, we rethought the process of generating multi-focus hash codes, which involves gradually fusing positional features into visual features. Therefore, we can save space by gradually remove the fused positional features through filtering to obtain hash codes with different focuses. This process can be described as:
\begin{equation}
	\begin{array}{c}
		\boldsymbol{b}_{(\tau-1)}^{'} = \left\{\begin{matrix} \boldsymbol{F} \cdot \boldsymbol{b}_{(\tau)}^{'}, \tau \ne \mathit{\Upsilon}
			\\ \boldsymbol{F} \cdot \boldsymbol{b}_{(\mathit{\Upsilon})}, \tau=  \mathit{\Upsilon}
		\end{matrix}\right. 
	\end{array}
\end{equation}
where $\boldsymbol{b}_{(\tau)}$ represents the layer $\tau$-th output original set of hash codes, and $\boldsymbol{b}'_{(\tau)}$ denotes the layer $\tau$-th compactly represented hash codes obtained through the filtering matrix $\boldsymbol{F}$. The approach reduces storage requirements; however, during the matrix multiplication process, the original binary codes may result in values exceeding 1. To address this, we first normalize the codes and then binarize them using the sign function. It can be formally described as:

\begin{equation}
	\boldsymbol{\widetilde{b}}_{(\tau)}=sign\left(\frac{\boldsymbol{b}'_{(\tau)}-\mu}{\sigma}\right),\tau \ne \mathit{\Upsilon}
\end{equation}
where $\boldsymbol{\widetilde{b}}_{(\tau)}$ represents the predict hash codes, $\mu$ and $\sigma$ denote the mean and standard deviation of the codes, respectively, and $sign()$ is the sign function that maps the normalized values to binary outputs. To achieve the retrieval effect of original hash code, when constructing the filtering matrix $\boldsymbol{F}$, we impose constraints through the MSE Loss to make $\boldsymbol{b}_{(\tau)}^{norm}$ as close as possible to $\boldsymbol{b}_{(\tau)}$. \\

This approach achieves significant storage savings by transforming the representation from $M \times \mathit{\Upsilon} \times K$ bits to $(M \times K + K^2)$ bits. The key advantage manifests as:
\begin{align}
	\text { CR }=\frac{M \times K + K^2}{M \times \mathit{\Upsilon} \times K} \notag
	\\	=\frac{1}{\mathit{\Upsilon}}+\frac{1}{M \cdot(\mathit{\Upsilon} / K)},
\end{align}
where CR is compression ratio. The storage overhead per video decreases as $M$ grows, approaching the theoretical limit of $1/\mathit{\Upsilon}$ compression; or practical scenarios where $M\gg K$, the ratio simplifies to $\approx 1/\mathit{\Upsilon}$, yielding consistent $\mathit{\Upsilon}$-fold savings. The space complexity evolves from $\mathcal{O}(M Y K)$ to $\mathcal{O}\left(M K+K^{2}\right)$.

\section{Experiment}\label{sec6}
\subsection{Experimental Setup}
No experiments are available for comparison as we are the first to propose the video activity hash problem. Using SVTH model, M-STVH model and group activity recognition datasets, we generate hash codes and then classify group activity recognition together to evaluate the performance. 

\subsubsection{Datasets}
We evaluated our approach on the following three public group activity recognition datasets that provide object tracking annotations and object action labels.

\textbf{Volleyball Dataset (VD)} \cite{VD}: The dataset comprises 4830 video clips extracted from 55 volleyball matches, with 3493 clips designated for training and 1337 for testing. Each clip contains 41 frames. Annotated within these clips are the coordinates of object bounding boxes, along with nine specific action labels (such as waiting, setting, digging, falling, spiking, blocking, jumping, moving, and standing) and eight group activity labels (including r-winpoint, r-passing, r-spiking, r-setting, l-setting, l-spiking, l-passing, and l-winpoint). We adhere to the actor coordinates and the training/testing partition outlined in references \cite{MLST,DIN,AFGf} to ensure comparability with group activity recognition methods.

\textbf{Collective Activity Dataset (CAD)} \cite{CAD}: The dataset comprises 44 video clips recorded using a low-resolution handheld camera, providing dynamic views. It encompasses five distinct collective activity labels (crossing, waiting, queueing, walking, talking), six individual action labels (NA, crossing, waiting, queueing, walking, talking), and eight individual posture labels (which are not utilized in our study). Group activity classes are determined based on the predominant actions observed within the video clip. We adopt a training/testing split of 2/3 for training and the remainder for testing to maintain consistency with prior experiments, as outlined in \cite{MLST}. Additionally, following \cite{DIN,SPTS, HiGCIN}, we consolidate the classes crossing and walking into a single class moving.

\textbf{Collective Activity Extended Dataset (CAED)} \cite{CAED}: The CAED dataset comprises 75 video clips, wherein two additional group activity categories, dancing and jogging, have been introduced compared to the CAD dataset. Moreover, the ambiguous activity class of walking has been eliminated due to its nature as more of an individual action rather than a group activity. We adopt the training/testing partition outlined in references \cite{MLST} and \cite{P2CTDM} to ensure equitable comparisons.

\begin{table*}[thp]
	\caption{The Predicted Results of Actions and Activities in VD, CAD, and CAED.' - ' Indicates that No Results are Provided(The best results are in bold)} \label{tab:tab1}
	\centering
	\begin{tabular}{@{}ccccccc@{}}
		
		\toprule
		\multirow{2}{*}{Method}       & \multirow{2}{*}{Modality} & \multirow{2}{*}{Bcakbone} & \multicolumn{2}{c}{VD}     & CAD           & CAED      \\ \cmidrule(l){4-7} 
		&                           &                           & Individual & Group         & Group         & Group                         \\ \hline
		Contextual Model\cite{ContextualModel}              & RGB                       & None                      & -          & -             & 83.4          & -                             \\
		Iterativae Belief ProPagation\cite{Itreative} & RGB                       & None                      & -          & -             & 79.0          & -                             \\
		HDTM \cite{VD}                         & RGB                       & AlexNet                   & -          & 81.9          & 81.5          & -                             \\
		SIM\cite{SIM}                           & RGB                       & AlexNet                   & -          & -             & 81.2          & 90.2                          \\
		RMIC \cite{RMIC}                         & RGB+Flow                  & Inception-v3              & -          & 66.9          & 86.1          & -                             \\
		SBGAR \cite{SBGAR}                        & RGB+Flow                  & AlexNet                   & -          & -             & 89.4          & -                             \\
		SPTS \cite{SPTS}                          & RGB+Flow                  & VGG16                     & -          & 91.2          & 95.8          & 98.1                          \\
		PMIC\cite{RMIC}                          & RGB                       & AlexNet                   & -          & 87.7          & 92.2          & -                             \\
		ARG  \cite{ARG}                         & RGB                       & Inception-v3              & 83.0       & 92.5          & 91.0          & -                             \\
		HiGCIN \cite{HiGCIN}                       & RGB                       & Resnet-18                 & -          & 91.5          & 93.4          & -                             \\
		DIN \cite{DIN}                          & RGB                       & VGG-16                    & -          & 93.6          & 95.9          & -                             \\
		P2CTDM  \cite{P2CTDM}                      & RGB                       & Inception-v3              & -          & 92.7          & 96.1          & 95.6                          \\
		stagNet \cite{StagNet}                      & RGB                       & VGG-16                    & 82.3       & 89.3          & 89.1          & 89.7                          \\
		CRM \cite{CRM}                          & RGB+FLOW                  & I3D                       & -          & 93.0          & 85.8          & -                             \\
		SAVRF \cite{SAVRF}                         & RGB+Flow                  & I3D                       & 83.1       & 95.0          & 95.2          & -                             \\
		Dual-AI \cite{dualAI}                      & RGB                       & Inception-v3              & 84.4       & 94.4          & 96.5          & -                             \\
		Actor-Transformer  \cite{Actorformer}           & RGB+Flow                  & I3D                       & 83.7       & 93.0          & 85.8          & -                             \\
		GroupFormer  \cite{Groupformer}                 & RGB                       & Inception-v3              & 83.7       & 94.1          & 96.5          & -                             \\
		MLST-Former  \cite{MLST}                 & RGB                       & Inception-v3              & 84.5       & 94.5          & 96.8          & 95.9                          \\
		RWGCN   \cite{RWGCN}                      & RGB                       & Resnet-50                 & -          & -             & 95.5          & 94.8                          \\
		AFGNet  \cite{AFGf}                      & RGB                       & Invception-v3             & 86.1       & \textbf{96.7} & 96.5          & 96.9                          \\
		MOGAR    \cite{MOGAR}                     & Keypoint                  & HRNet                     & -          & 94.5          & 94.4          & -                             \\ \hline
		STVH(Ours)                    & RGB                       & VGG-16                    & 87.6       & 95.6          & \textbf{98.3} & \textbf{99.0}  \\
		M-STVH(Ours)                    & RGB                       & Invception-v3             & \textbf{87.7}          & 95.7          & 97.8 &  98.5                           \\ \hline
	\end{tabular}
\end{table*}

\subsubsection{Experimental Setting}
We employ the Inception-v3 model \cite{inv3} pre-trained on ImageNet \cite{imagenet} as the CNN backbone network to ensure consistency with methods such as \cite{SPTS}, enabling fair comparisons. For the training and testing on the VD dataset, we set $T=10$ and the video frame resolution to $H\times W=720\times1280$. Conversely, for the CAD and CAED datasets, we also set $T=10$ but adjust the frame resolution to $H\times W=480\times720$. Additionally, we specify the number of objects in group activities as $N=12$ for the VD dataset and $N=13$ for the CAD and CAED datasets.  For group behavior retrieval, we use the training set as the base set and the testing set as the query set. For group visual analysis, we mix all videos and use video IDs as labels to evaluate the effectiveness of multi-focused video hashing retrieval. Object features are extracted using RoIAlign, which facilitates cropping and resizing to a fixed size of $5 \times 5$. Optimization across all datasets is performed using Adam, with the learning rate configured as follows: for the VD dataset, the initial learning rate is set to $1\times 10^{-5} $, decaying to $5\times 10^{-6} $ at the 11th epoch and further to $1\times 10^{-6} $ at the 21st epoch. For the CAD and CAED datasets, the learning rate remains constant at $1\times 10^{-5} $ throughout the training process without decay. All datasets are trained for 60 epochs in STVH, with $\lambda_1$=0.1 and $\lambda_1$=0.5 applied uniformly. All datasets are trained for 60 epochs in M-STVH, with $\mu_1$=0.1, $\mu_1$=0.5, and $\mu_3$=0.01 applied uniformly. All experiments are conducted using PyTorch 1.10 with CUDA 11.3, a batch size of 2, and executed on a machine equipped with 2 GTX 4090 GPUs.

\subsubsection{Evaluation Metrics}
Precision is a basic metric used to assess classification accuracy, which indicates the similarity between the predicted labels and the ground truth by calculating TP/(TP + FP). TP is the number of samples in which the positive predicted labels are the same as the ground truth, and FP is the number of samples in which the positive prediction is incorrect. This metric can be used to assess the reliability of model classification effectively.

We assessed the retrieval accuracy of the STVH using the mean average precision (mAP). We first calculate AP@k, which denotes the average precision of the top [1..k] results for each sample retrieval, to compute mAP@k. Subsequently, mAP@k is derived by averaging all the AP@k scores. This metric provides a comprehensive evaluation of the system's precision, particularly focusing on the effectiveness of returning relevant results within the top k items for each query or instance.

\subsection{Group Activity Classification }
In this section, STVH and M-STVH  compare with state-of-the-art methods on three datasets: VD, CAD, and CAED, respectively.  Previous methods focus solely on group activity recognition, while STVH not only generates efficient hash codes but also simultaneously performs activity classification tasks.  Furthermore, M-STVH enhances the model's expressive capability by generating hash codes that focus on different semantic information.  In this section, we perform activity classification based on the hash codes generated by M-STVH that focus on group activity information to validate its effectiveness in distinguishing between different activities.

As shown in Table 2, our method demonstrates strong competitiveness across multiple datasets. On the VD dataset, our approach outperforms other methods using VGG16 backbone, such as StagNet \cite{StagNet} and DIN \cite{DIN}, by nearly 3\% in group activity classification. Compared with methods using Inception-v3 as the backbone network, such as MLST-Former \cite{MLST} and GroupFormer \cite{Groupformer}, our method has also improved by nearly 2\% in the classification of group activities, lagging only slightly behind AFGNet \cite{AFGf} by 1\%. Nevertheless, our method performs classification while generating hash codes. In particular, M-STVH needs to generate multi-focused hash encoding. However, regardless of the backbone network used for feature extraction, our method achieves the best results in individual action classification. Meanwhile, on both CAD and CAED datasets, our method achieves at least 3\% improvement in group activity classification accuracy compared to other state-of-the-art approaches, including AFGNet \cite{AFGf}, RWGCN \cite{RWGCN}, and MLST-Former \cite{MLST}. Furthermore, after generating multi-focused hash encoding using M-STVH, there is no decrease in classification accuracy compared to STVH. This highlights the effectiveness of the MSF module in performing spatiotemporal interleaving at a high semantic level.

\begin{figure*}[h]
	\centering  
	\subfloat[Confusion matrix of group activity prediction on\\ VD using STVH]{
		\label{fig_VD_STVH}\includegraphics[width=0.5\textwidth]{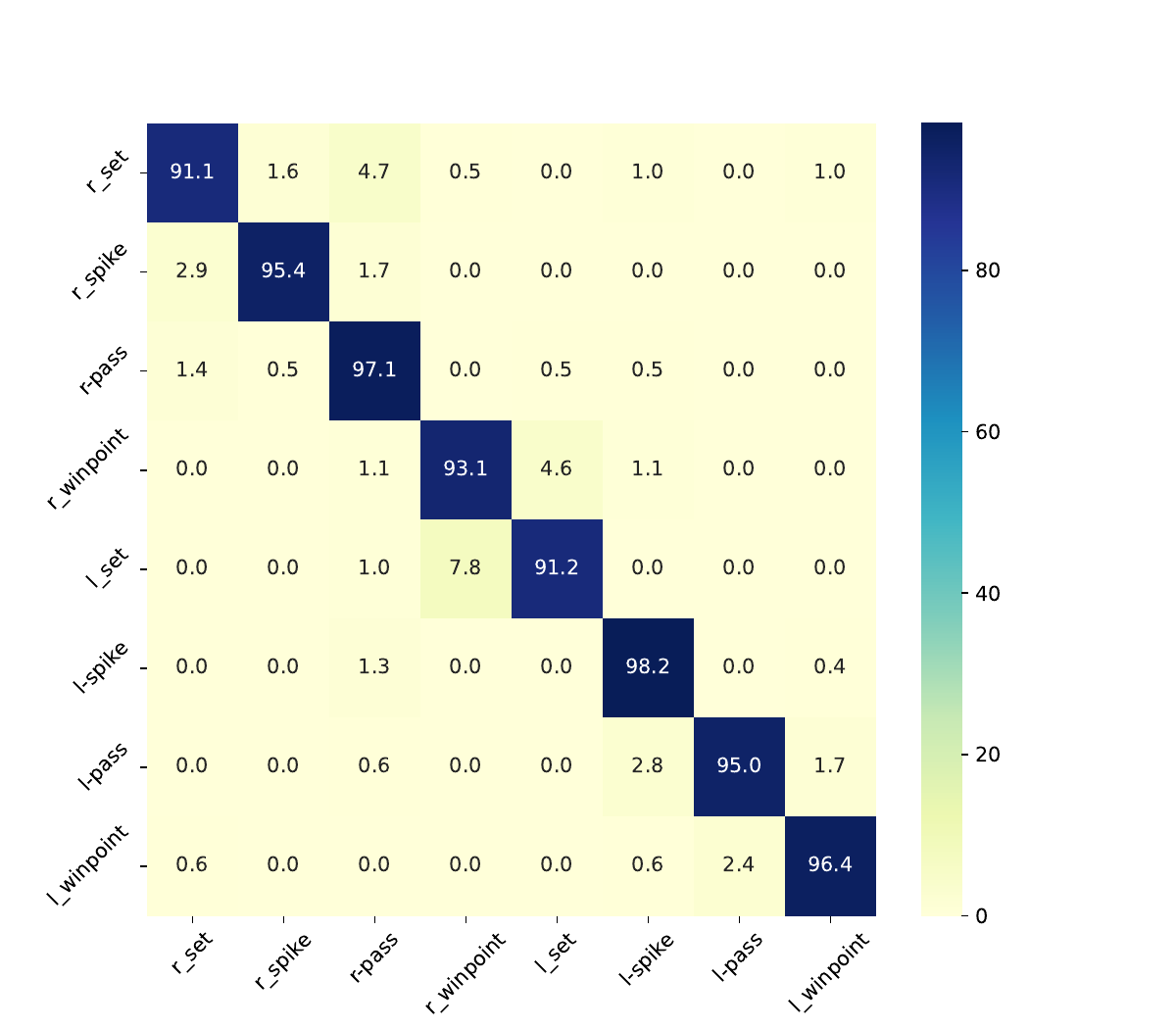}
	} 
	\subfloat[Confusion matrix of group activity prediction on \\VD using M-STVH]{
		\label{fig_VD_M-STVH}\includegraphics[width=0.5\textwidth]{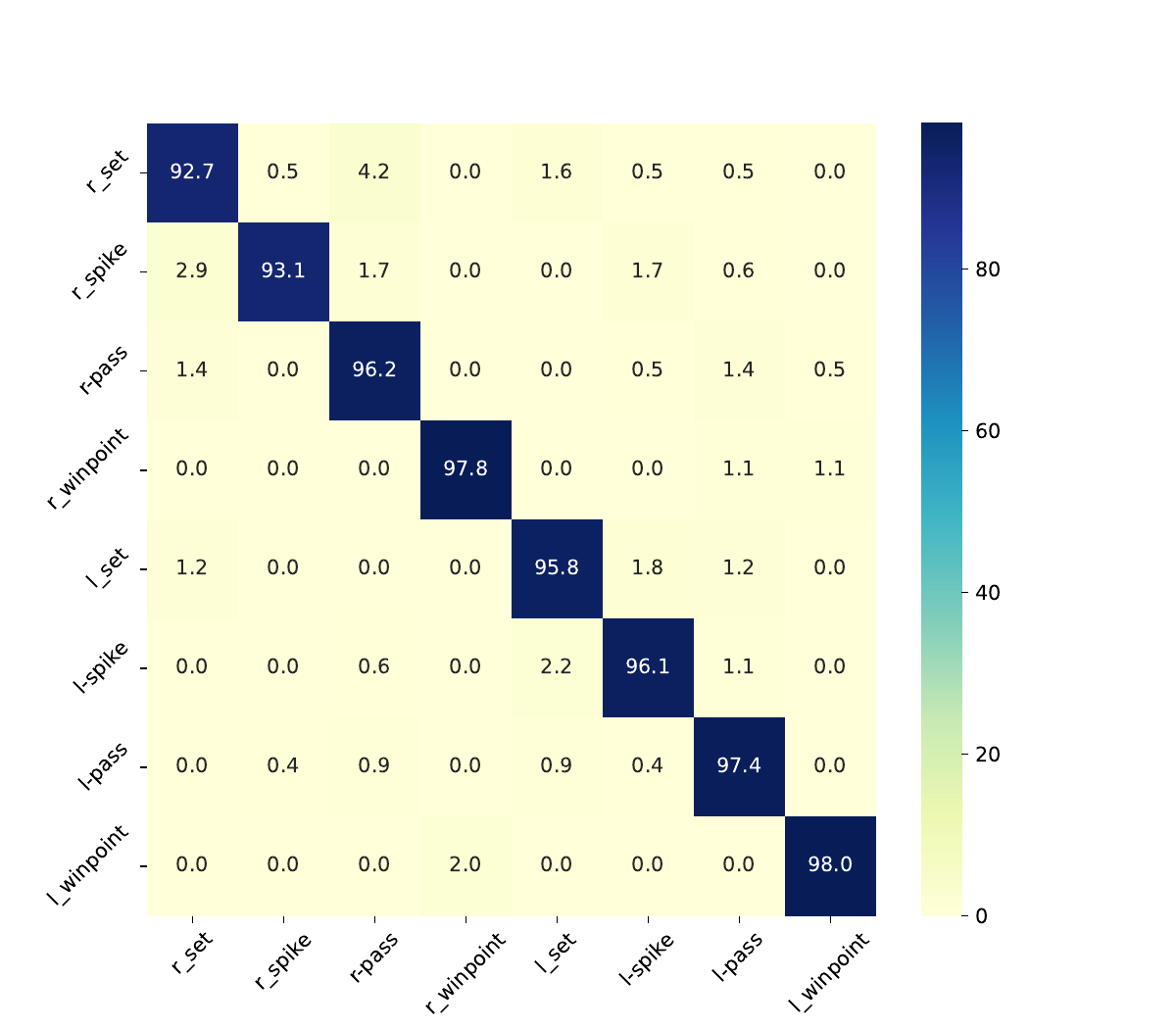}
	}            
	
	\caption{Confusion matrix of group activity prediction on the VD, where vertical axis represent predicted labels and horizontal axis indicate ground truth}   
	\label{VD_Confusion}
\end{figure*}

\begin{figure*}[h]
	\centering  
	\subfloat[Confusion matrix of group activity prediction on\\ CAD using STVH]{
		\label{fig_CAD_STVH}\includegraphics[width=0.5\textwidth]{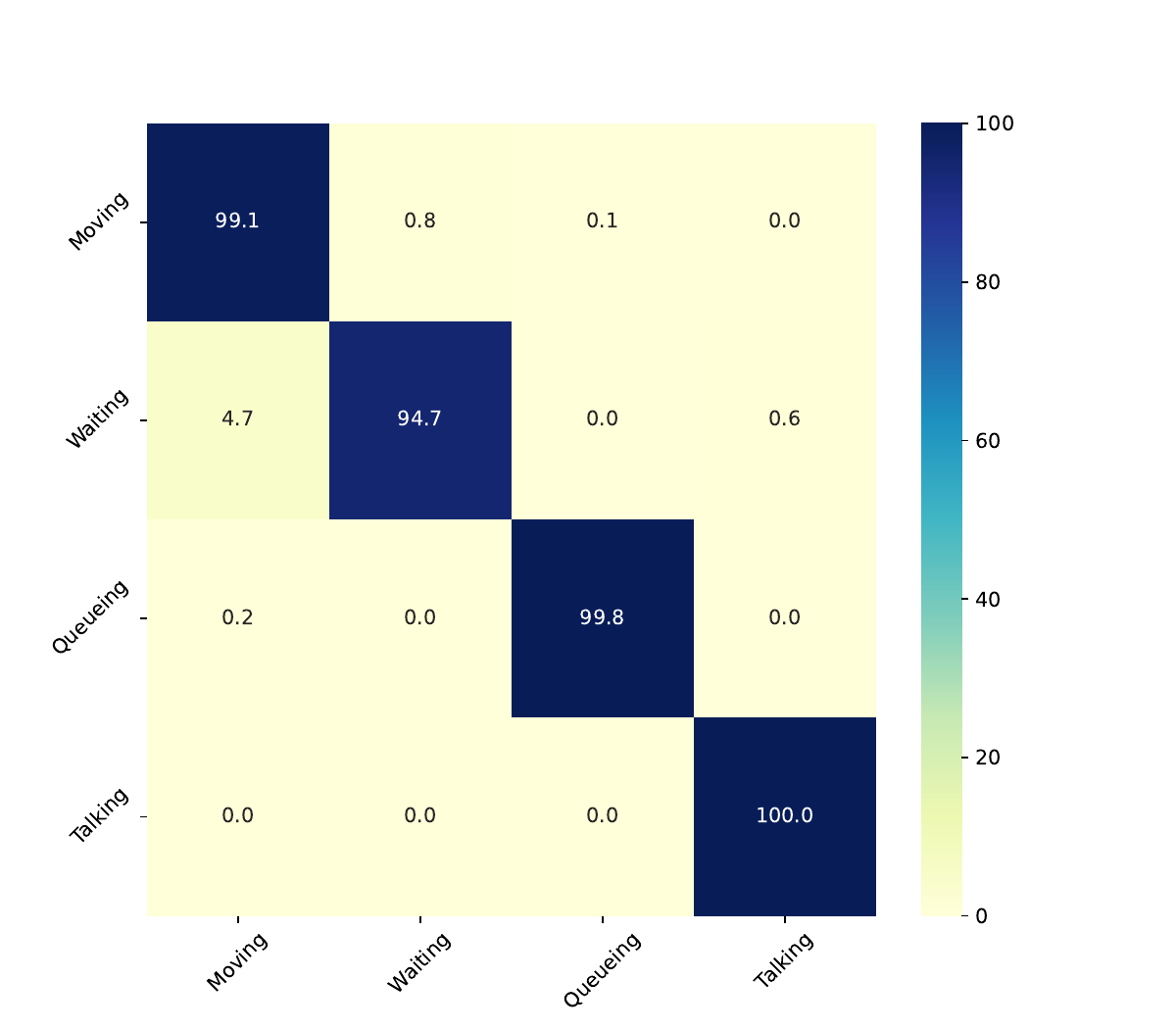}
	} 
	\subfloat[Confusion matrix of group activity prediction on \\CAD using M-STVH]{
		\label{fig_CAD_M-STVH}\includegraphics[width=0.5\textwidth]{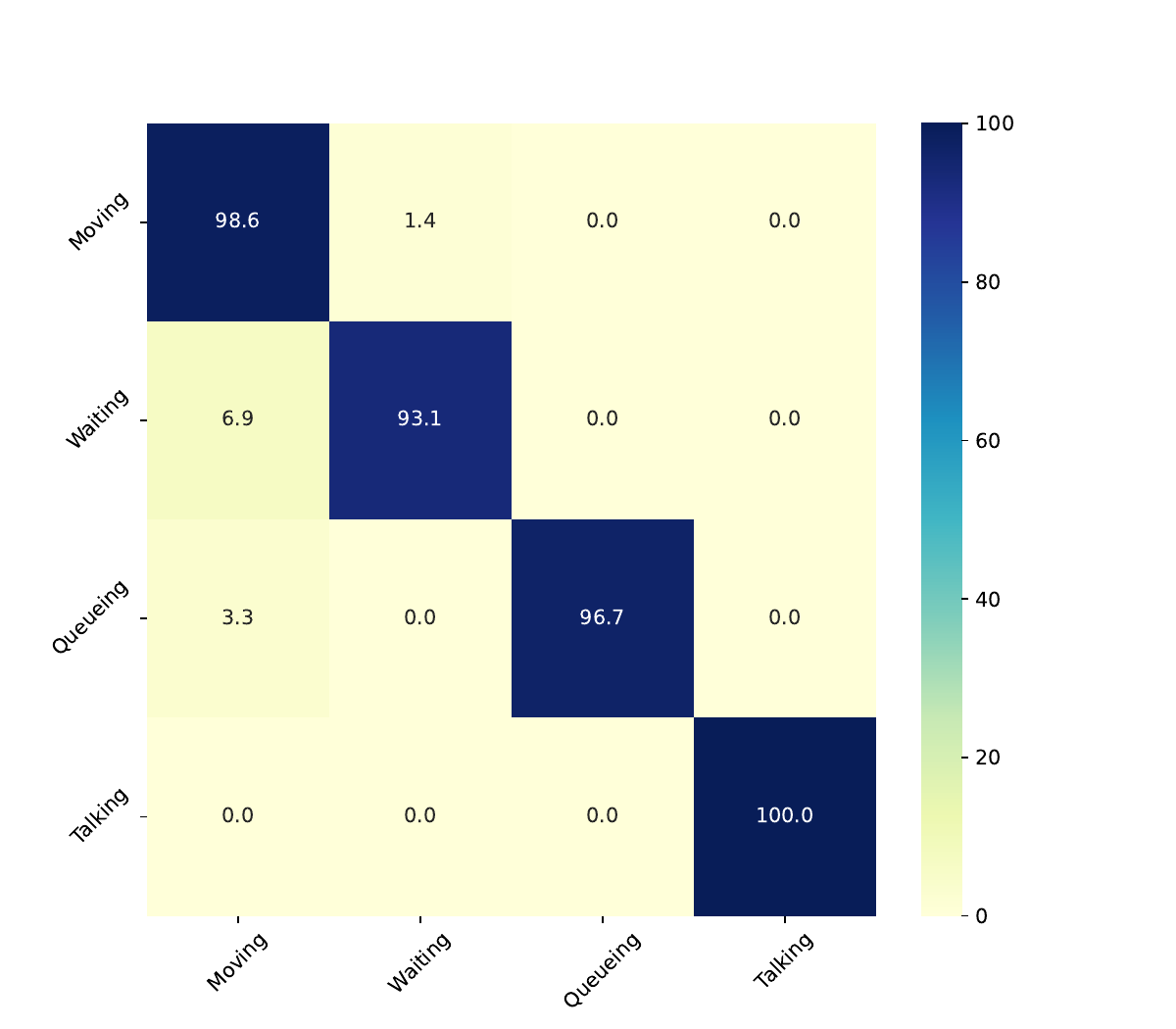}
	} \\           
	\caption{Confusion matrix of group activity prediction on the CAD, where vertical axis represent predicted labels and horizontal axis indicate ground truth}   
	\label{CAD_and_CAED_Confusion}
\end{figure*}

As shown in Figs. \ref{VD_Confusion} and \ref{CAD_and_CAED_Confusion}, we compared the confusion matrices of STVH and M-STVH for group activity classification on the VD and CAD datasets. STVH focuses on features of the group activity itself, whereas M-STVH simultaneously attend to the multi-focus hashing for visual features and group activities. Consequently, STVH generally outperforms M-STVH in group activity classification tasks. However, due to strongly symmetrical visual features of objects on the VD dataset, it effectively assists M-STVH during multi-focus hashing learning and the promotes activity classification tasks. 

\subsection{Group Activity Retrieval}
To validate the accuracy of the STVH and M-STVH group activity retrieval framework, we conduct comprehensive experiments using the training sets from both the VD and CAD datasets as the database, while employing their  test sets as query sets. The number of stacking layers of PVF in STVH and MSF in M-STVH are the same. Since STVH can only retrieve videos at the group activity, in this experiment, we only use the output hash encoding of the last layer of M-STVH for comparison.

\begin{table*}[]
	\centering
	\caption{Retrieved mAP on the Volleyball Dataset (best results in bold)}
	\label{tab:tab2}
	\begin{tabular}{@{}ccccccccc@{}}
		\toprule
		\multirow{2}{*}{} & \multicolumn{2}{c}{16bits} & \multicolumn{2}{c}{32bits} & \multicolumn{2}{c}{64bits} & \multicolumn{2}{c}{128bits}     \\ \cmidrule(l){2-9} 
		& STVH        & M-STVH       & STVH        & M-STVH       & STVH        & M-STVH       & STVH           & M-STVH         \\ \midrule
		mAP@5             & 94.23       & 94.93        & 95.11       & 95.72        & 95.07       & 95.85        & 95.99          & \textbf{96.12} \\
		mAP@10             & 93.66       & 94.42        & 95.10       & 95.23        & 95.07       & 95.47        & \textbf{96.02} & 96.00          \\
		mAP@20             & 93.82       & 94.33        & 95.14       & 94.54        & 95.04       & 94.87        & \textbf{95.67} & 95.66          \\
		mAP@50             & 94.20       & 94.55        & 95.09       & 94.59        & 95.03       & 94.62        & \textbf{95.68} & 95.58          \\ \midrule
		Classification    & 94.84       & 95.06        & 94.99       & 95.21        & 95.21       & 95.59        & 95.59          & \textbf{95.66} \\ \bottomrule
	\end{tabular}
\end{table*}

\begin{table*}[]
	\centering
	\caption{Retrieved mAP on the Collective Activity Dataset (best results in bold)} \label{tab:tab3}
	\begin{tabular}{@{}ccccccccc@{}}
		\toprule
		\multirow{2}{*}{} & \multicolumn{2}{c}{16bits} & \multicolumn{2}{c}{32bits} & \multicolumn{2}{c}{64bits} & \multicolumn{2}{c}{128bits} \\ \cmidrule(l){2-9} 
		& STVH        & M-STVH       & STVH        & M-STVH       & STVH        & M-STVH       & STVH             & M-STVH   \\ \midrule
		mAP@5             & 97.17       & 91.83        & 97.07       & 96.49        & 97.26       & 96.28        & \textbf{98.24}   & 96.91    \\
		mAP@10             & 97.73       & 93.64        & 97.04       & 96.30        & 97.28       & 96.56        & \textbf{98.10}   & 96.81    \\
		mAP@20             & 97.69       & 94.67        & 97.22       & 96.15        & 96.79       & 96.89        & \textbf{98.10}   & 96.78    \\
		mAP@50             & 97.80       & 94.99        & 97.21       & 96.53        & 96.33       & 96.91        & \textbf{97.91}   & 96.82    \\ \midrule
		Classification    & 97.76       & 95.82        & 97.37       & 96.99        & 97.89       & 96.99        & \textbf{98.29}   & 97.78    \\ \bottomrule
	\end{tabular}
\end{table*}

As shown in Tables \ref{tab:tab2} and \ref{tab:tab3}, STVH demonstrates superior retrieval performance when performing group activity retrieval at different hash lengths. The results are consistent with expectations, as M-STVH requires the simultaneous generation of multiple attention hash encodings, which theoretically increases retrieval complexity. 

\subsection{Multi-Focused Group Activity Retrieval}
Focus on visual retrieval evaluation, we introduce a novel grouping criterion specific to the VD dataset: videos from the same volleyball match are considered to share identical group visual features.  The assumption holds because players maintain consistent uniforms for a single match, providing a reliable ground truth for visual similarity assessment.

\begin{figure*}[]
	\centerline{\includegraphics[width=16cm]{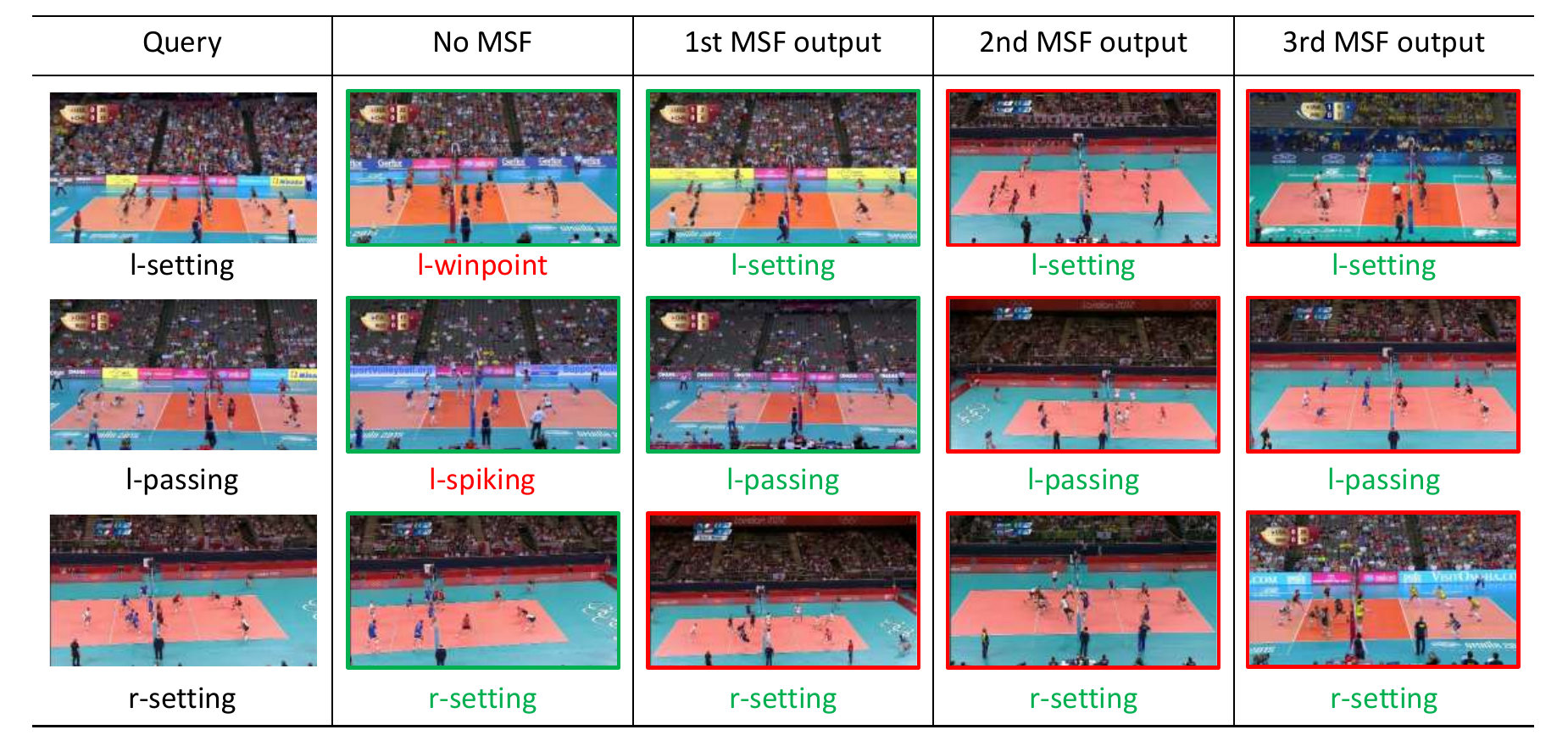}}
	\caption{The top-1 video in multi-focus hash retrieval, where the image displayed in a green box indicates that the query sample and the retrieved sample have the same visual characteristics. In contrast, the red box indicates that they are different. Green font indicates that the query sample and the retrieved sample have the same activity category, while red indicates that they are different.}
	\label{max_sim}
\end{figure*}

As shown in Fig. \ref{max_sim}, the top 1 retrieval results on different focuses provide strong evidence for the multi-focued representation capability. For the initial hash codes without MSF processing, the most similar retrieved samples primarily share identical visual features. However, as we utilize hash codes with progressively deeper MSF layers, the retrieval focus undergoes a remarkable transition from visual-based to activity-based retrieval, while preserving reasonable visual relevance. The transition conclusively demonstrates M-STVH's effectiveness in learning hierarchical representations that adaptively balance visual and activity information. The quantitative consistency between these visual results and our earlier mAP measurements further validates the robustness of our approach.

\begin{figure}[]
	\centering  
	\subfloat[Map@5]{
		\label{fig8_a}\includegraphics[width=0.4\textwidth]{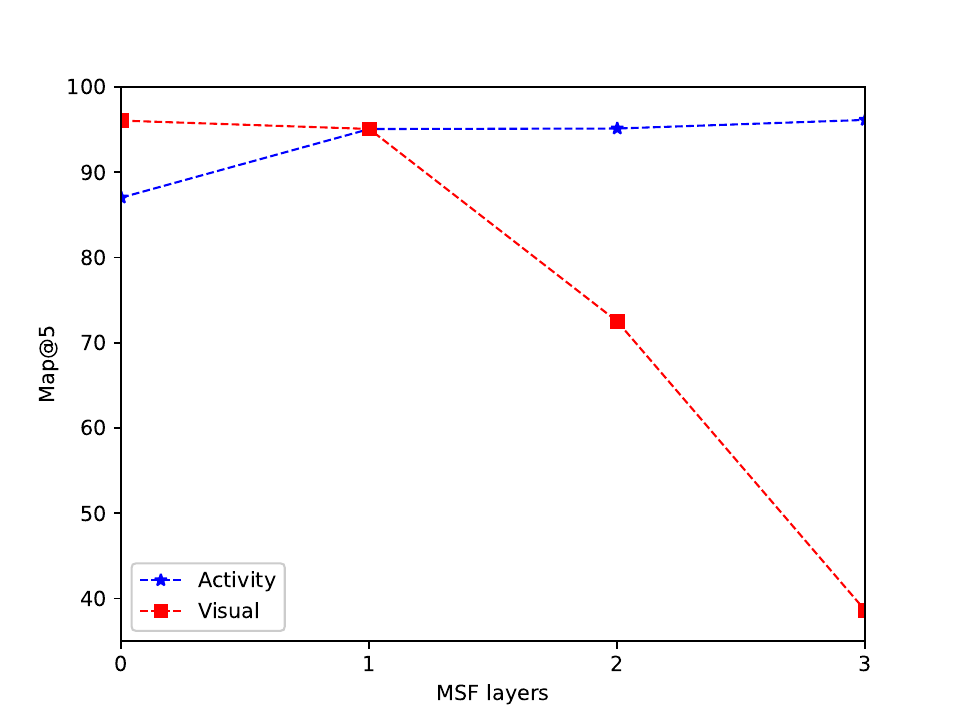}
	}
	\subfloat[Map@10]{
		\label{fig8_b}\includegraphics[width=0.4\textwidth]{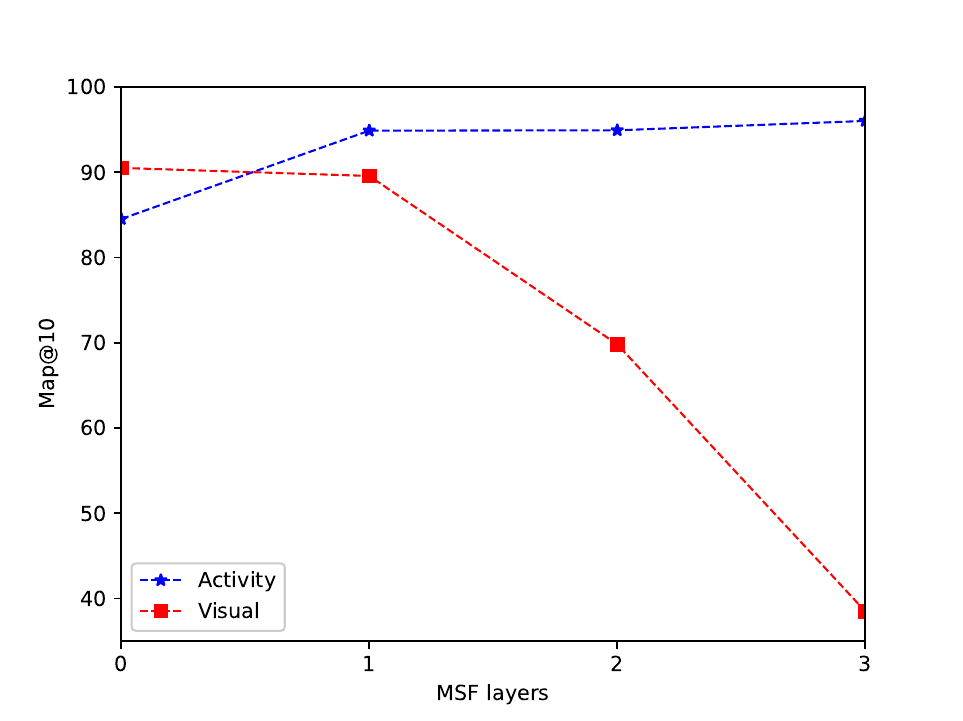}
	}          
	
	\caption{Hierarchical retrieval performance (mAP) on VD dataset with 128-bit hash codes, evaluating both group appearance and group activity retrieval across different MSF output layers (horizontal axis)}   
	\label{MSF_}
\end{figure}

Fig. \ref{MSF_} shows the hash retrieval accuracy of M-STVH in group vision and group activities with the number of layers of the MSF module stack. The results show that as the number of MSF stacking layers increases, the retrieval effect of group activities has significantly improved, and the retrieval effect of group vision has also decreased. These findings provide strong empirical evidence for M-STVH's effectiveness in learning disentangled yet complementary representations, where shallow layers capture visual appearance patterns and deeper layers encode sophisticated group interactions, ultimately enabling comprehensive multi-focused video understanding.  

\begin{table*}[]
	\centering
	\caption{Classification accuracy and mAP@10 at 128 bits after replacing the MSF module} \label{tab:tab5}
	\begin{tabular}{@{}ccccccc@{}}
		\toprule
		\multirow{2}{*}{Dataset} & \multicolumn{2}{c}{$\text{M-STVH}^{\text{BD}}$} & \multicolumn{2}{c}{$\text{M-STVH}^{\text{ED}}$} & \multicolumn{2}{c}{M-STVH}        \\
		& precision        & mAP@10       & precision        & mAP@10       & precision      & mAP@10         \\ \midrule
		VD                       & 94.69            & 93.93        & 95.21            & 94.56        & \textbf{95.66} & \textbf{96.00} \\
		CAD                      & 96.32            & 96.77        & 95.53            & 95.36        & \textbf{97.78} & \textbf{96.81} \\
		CAED                     & 94.25            & 90.26        & 97.35            & 96.05        & \textbf{98.50} & \textbf{98.00} \\ \bottomrule
	\end{tabular}
\end{table*}

\subsection{Ablation Experiments}
We conducted multiple sets of ablation experiments on M-STVH from the perspectives of both its model architecture and loss functions, thereby confirming the effectiveness of the proposed modules.

\subsubsection{Model Architecture Ablation Experiments}
In terms of integrating visual and positional features, we replaced the MSF module with a transformer module to validate the effectiveness of the MSF module. Initially, we substituted the MSF module with a traditional transformer block. To ensure the integration of positional features during training, we then added positional features to the input ($\text{M-STVH}^{\text{BD}}$) and output ($\text{M-STVH}^{\text{ED}}$) of the transformer block. Table \ref{tab:tab5} shows that the MSF module improved the final classification results and achieved the best experimental outcomes. It demonstrates that when modeling the activities of multiple objects in a video, the transformer block may encounter bottlenecks. Similarly, the relatively limited results obtained by simply summing positional and visual information at the feature indicate the effectiveness of MSF in fusing positional information and visual information at a higher semantic level.

\begin{table*}[]
	\centering
	\caption{Classification Accuracy and mAP@10 at 128 bits with Different Numbers of MFS Module Stacked Layers}
	\label{tab:tab6}
	\begin{tabular}{@{}ccccccccc@{}}
		\toprule
		\multirow{2}{*}{Dataset} & \multicolumn{2}{c}{$\mathit{\Upsilon } =2$} & \multicolumn{2}{c}{$\mathit{\Upsilon } =3$} & \multicolumn{2}{c}{$\mathit{\Upsilon } =4$} & \multicolumn{2}{c}{$\mathit{\Upsilon } =5$} \\ \cmidrule(l){2-9} 
		& precision              & mAP@10             & precision            & mAP@10               & precision            & mAP@10               & precision              & mAP@10             \\ \midrule
		VD                       & 93.87                  & 93.49                   & \textbf{95.66}       & \textbf{96.00}       & 94.99                & 95.34            &  94.39                    &  95.05                  \\
		CAD                      & 96.21                    & 88.91          &  96.74                    & 95.87                     & \textbf{97.78}       & \textbf{96.81}       &  95.29                      &  95.37                  \\
		CAED                     & 97.01                  & 95.87              & 97.67                & 96.50                & \textbf{98.50}       & \textbf{98.00}       & 97.12                  & 95.12              \\ \bottomrule
	\end{tabular}
\end{table*}

The experiments show progressive fusion approach of the MSF module outperforms direct addition. To validate the impact of different numbers of stacked layers within the MSF module on the fusion of visual and positional features, we varied the stack layers $\mathit{\Upsilon }$ = 2, 3, 4, and 5. The specific results are outlined in Table \ref{tab:tab6}. Notably, the optimal experimental outcomes are achieved with $\mathit{\Upsilon } =3$ on the VD dataset, while in the CAD and CAED datasets, the best results emerge with  $\mathit{\Upsilon } =4$. It is attributed to the heightened importance of visual features in volleyball matches compared to those in the CAD and CAED datasets. This experiment underscores that as the number of stacked layers in the MSF module increases, STVH increasingly emphasizes the positional features of groups rather than visual features.

\begin{figure*}[t]   
	\centering           
	\subfloat[]  
	{
		\label{fig10_VD}\includegraphics[width=0.95\textwidth]{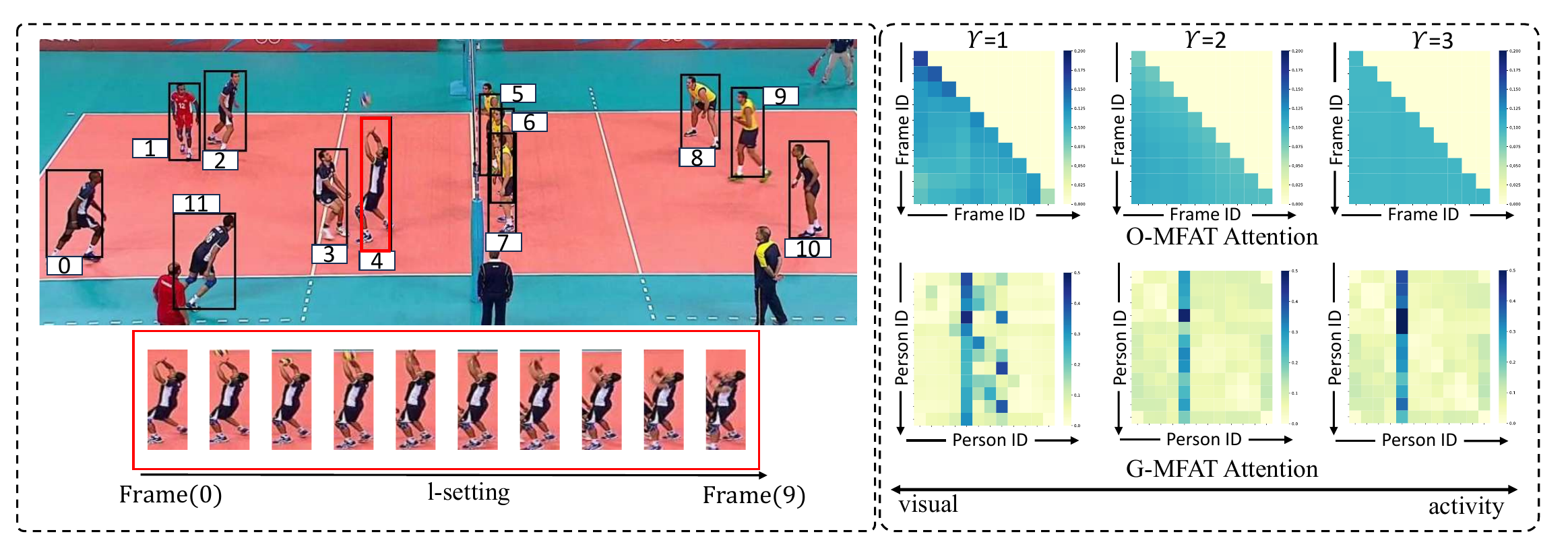}
	} \\
	\subfloat[]
	{
		\label{fig10_CAD}\includegraphics[width=0.95\textwidth]{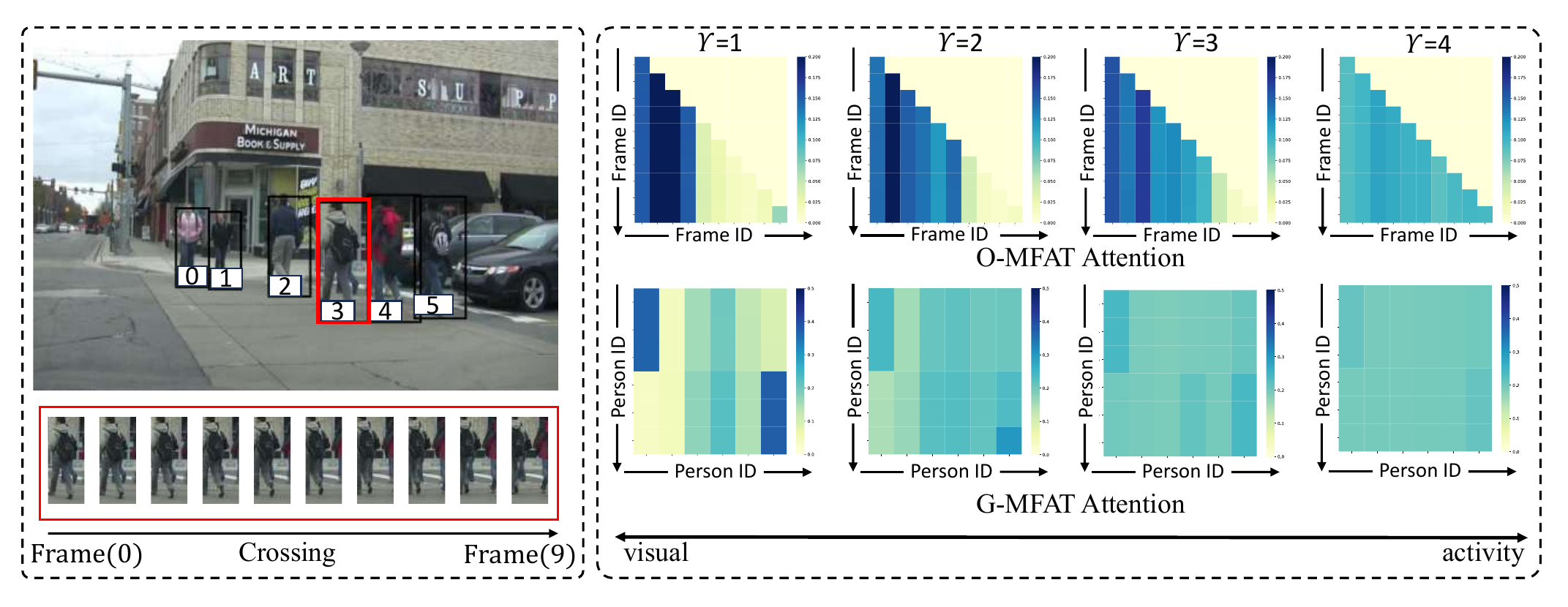}
	}
	\caption{In group activity inference, the attention matrices on different layers of the MSF. (a) results on VD; (b) results on CAD.}    
	\label{fig10}
\end{figure*}

During the inference process of M-STVH, we visualized the attention matrices across layers to conduct an in-depth analysis of the attention distribution mechanisms employed by O-MFAT and G-MFAT during MSF iterative computations. In shallow layers in Fig. \ref{fig10}, the attention matrix of the O-MFAT places emphasis on specific frames as visual features exhibit no significant variation within the video segment. With the number of layers increases, O-MFAT expands its scope of attention from the initial few frames to all video frames, because all the frames may contribute to the action recognition. Similarly, at shallow layers, due to insufficient integration of positional information, G-MFAT exhibits vague modelling of object interactions, with the attention matrix displaying considerable disorder. As the number of network layers increases, G-MFAT exhibits two entirely distinct activity in VD and CAD. As shown in Fig. \ref{fig10_VD}, it focuses on the person who play pivotal roles in group activities, while in Fig. \ref{fig10_CAD}, it focuses on all the persons across the entire group. The divergence in attention evolution stems from different definitions of group activity across the two datasets: in the VD dataset, group activity labels were primarily determined based on the behaviour of the key person; whereas in the CAD dataset, group activity categories are often determined by the joint actions of all persons.

\subsubsection{Loss Function Ablation Experiments}
\begin{table*}[h]
	\centering
	\caption{Classification accuracy and mAP@10 at 128 bits with different loss function constraints}
	\label{tab:tab7}
	\begin{tabular}{@{}ccccccc@{}}
		\toprule
		\multirow{2}{*}{Dataset} & \multicolumn{2}{c}{$L_{cls}+{\mu}_3 L_{recon}$} & \multicolumn{2}{c}{$L_{cls} + \mu_1L_q+{\mu}_3 L_{recon}$} & \multicolumn{2}{c}{$L_{cls} + \mu_1L_q + \mu_2L_H+{\mu}_3 L_{recon}$} \\ \cmidrule(l){2-7} 
		& precision                  & mAP@10                 & precision                        & mAP@10                          & precision                               & mAP@10                                  \\ \midrule
		VD                       & 94.24                           &   94.72                     & 94.37                       &  94.86                     & \textbf{95.66}                          & \textbf{96.00}                          \\
		CAD                      & 97.50                           &  95.50                      & 96.97                                 &97.39                                 & \textbf{97.78}                          & \textbf{96.81}                          \\
		CAED                     & 97.35                           &  91.91                      & 97.46                                 &  94.63                               & \textbf{98.50}                          & \textbf{98.00}                          \\ \bottomrule
	\end{tabular}
\end{table*}

We experimented with three different combinations of loss functions, $L_{cls}+\mu_3 L_{recon}$, $L_{cls}+ \mu_1L_q+\mu_3 L_{recon}$, and $L_{cls}+ \mu_1L_q+ \mu_2L_H+\mu_3 L_{recon}$, using accuracy and mAP@10 metrics to analyze the impact of each component of the total loss function on the results. Table \ref{tab:tab7} shows that when training solely with  $L_{cls}$, the M-STVH achieved good classification results on several datasets, but its retrieval performance is not satisfactory. Upon incorporating $\mu_1L_q$, the retrieval performance improved, albeit with a slight decrease in classification accuracy. With the addition of $\mu_2L_H$, the retrieval results further improved, and the classification accuracy increased, achieving the best performance. The findings validate the positive role played by each designed loss function in the training process of the M-STVH.

\begin{figure}[h]   
	\centering           
	\subfloat[With hash loss]  
	{
		\label{hash_loss}\includegraphics[width=0.4\textwidth]{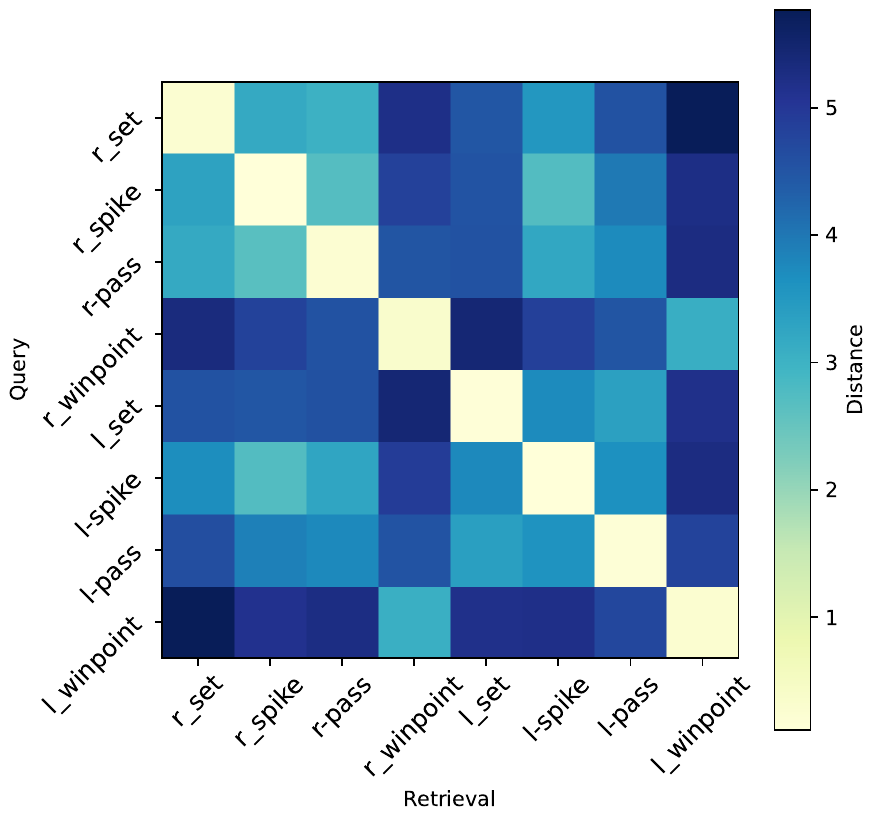}
	} 
	\subfloat[Without hash loss]
	{
		\label{without_hashloss}\includegraphics[width=0.4\textwidth]{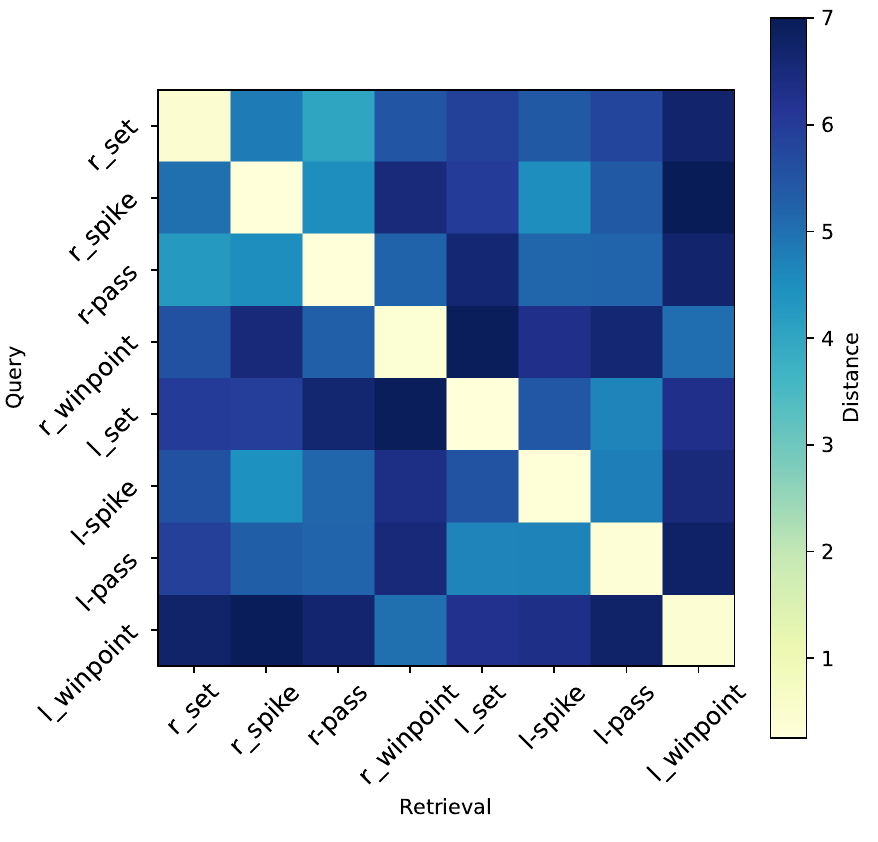}
		
	}
	\caption{The average Hamming distance between different class hash codes in a VD dataset at 128 bits.}    
	\label{hashloss}
\end{figure}

As shown in Fig. \ref{hashloss}, the comparison of average Hamming distances between hash codes reveals the effectiveness of our proposed contrastive loss. The results demonstrate that our method maintains sufficiently large Hamming distances between fundamentally different action categories (e.g.,``l-winpoint" and ``r-set"), while appropriately reducing the distance between semantically similar actions (e.g., ``r-set", ``r-spike", and ``r-pass"). The pattern indicates that the contrastive loss based on object distribution similarity successfully preserves the discriminability of dissimilar categories while bringing closer those with similar group activity patterns. The Hamming distance distribution across categories further confirms that our approach organizes the hash space in a semantically meaningful way, where the relative distances between different types of group activities correspond well to their actual behavioral relationships in volleyball games. These observations validate that the learned hash codes effectively capture both the distinctions and similarities between different group activities.

\section{Conclusion} \label{sec7}
This paper presents a novel solution for group activity retrieval through two key contributions: the spatiotemporal video Hashing (STVH) model and its enhanced version, multi-focused spatiotemporal video hashing (M-STVH). The STVH framework pioneers a dual spatiotemporal interleaving approach that effectively models group activities by jointly analyzing object dynamics and group interactions, capturing both visual feature evolution and positional relationships. Building upon this foundation, M-STVH introduces a hierarchical multi-step fusion mechanism, which enables progressive integration of visual and positional features across multiple representation layers. The innovative architecture naturally transitions from shallow-layer visual feature extraction to deep-layer activity semantics understanding, while the incorporated binary filtering matrix significantly optimizes storage efficiency. The proposed method not only advances multi-focused semantic modeling in video analysis but also demonstrates substantial practical potential for applications ranging from sports analytics to intelligent surveillance systems. In future we will explore cross-camera correlation analysis to further extend the framework's capability for large-scale

\begin{acks}
	This work was supported in part by China NSF Grant No.62271274, Ningbo S\&K Project Grant No.2024Z004, No.2023Z059, Zhejiang Provincial Natural Science Foundation of China under Grant No.ZCLQN25F0207, and the programs sponsored by K. C. Wong Magna Fund in Ningbo University.
\end{acks}
\bibliographystyle{ACM-Reference-Format}
\bibliography{sample-base}


\end{document}